\documentclass{article}

\usepackage{amsmath,amssymb,amsthm}
\usepackage{bbm}
\usepackage{bm}
\usepackage{balance}
\usepackage{booktabs} % for professional tables
\usepackage{enumitem}
\usepackage[T1]{fontenc}
\usepackage{graphicx}
\usepackage{nicefrac}
\usepackage{mathtools}
\usepackage{microtype}      % microtypography
\usepackage{physics}
\usepackage{subfigure}
\usepackage{xcolor}
\usepackage{wrapfig}
\definecolor{C0}{HTML}{1f77b4}
\definecolor{C1}{HTML}{ff7f0e}
\definecolor{C2}{HTML}{2ca02c}
\definecolor{C3}{HTML}{d62728}
\definecolor{C4}{HTML}{9467bd}
\definecolor{C5}{HTML}{8c564b}
\definecolor{C6}{HTML}{e377c2}
\definecolor{C7}{HTML}{7f7f7f}
\definecolor{C8}{HTML}{bcbd22}
\definecolor{C9}{HTML}{17becf}

\DeclareMathOperator{\cov}{\mathsf{Cov}}

\DeclareMathOperator{\pmse}{\sf pmse}

\DeclareMathOperator{\snr}{\sf snr}

\newcommand{\sqD}{\sqrt{D}}
\newcommand{\sqP}{\sqrt{P}}

\newcommand{\eclass}[1]{\epsilon_{c#1}}
\newcommand{\ussD}{\frac{1}{\sqrt{D}}}

\renewcommand{\dd}{\operatorname{d}\!}
\newcommand{\EE}{\mathbb{E}\,}

\renewcommand{\erf}{\mathrm{erf}}

\newcommand{\reals}{\mathbb{R}}

\newcommand{\normal}{\mathcal{N}}

\usepackage{multicol}
\usepackage{balance}

\newcommand{\textcite}[1]{\citet{#1}}

\usepackage{hyperref}
 \usepackage[accepted]{icml2021}

\icmltitlerunning{Kernels vs neural networks for Gaussian mixture classification}

\begin{document}

\twocolumn[ \icmltitle{Classifying high-dimensional Gaussian mixtures:\\ Where kernel methods fail and neural networks succeed}



\icmlsetsymbol{equal}{*}

\begin{icmlauthorlist}
\icmlauthor{Maria Refinetti}{ens,ide}
\icmlauthor{Sebastian Goldt}{sissa}
\icmlauthor{Florent Krzakala}{ide}
\icmlauthor{Lenka Zdeborová}{spo}
\end{icmlauthorlist}

\icmlaffiliation{spo}{SPOC Lab, EPFL}
\icmlaffiliation{ens}{Laboratoire de Physique de l’École Normale Supérieure, Université PSL, CNRS, Sorbonne Université, Université Paris-Diderot, Sorbonne Paris Cité}
\icmlaffiliation{sissa}{International School of Advanced Studies (SISSA), Trieste, Italy}
\icmlaffiliation{ide}{IdePHICS Lab, EPFL}

\icmlcorrespondingauthor{}{maria.refinetti@ens.fr}

\icmlkeywords{  Gaussian mixture classification, neural networks, random features, stochastic
  gradient descent.}

\vskip 0.3in
]



\printAffiliationsAndNotice{} % otherwise use the standard text.

% 4-6 sentences 
\begin{abstract}
A recent series of theoretical works showed that the dynamics of neural
  networks with a certain initialisation are well-captured by kernel methods.
  Concurrent empirical work demonstrated that kernel methods can come close to
  the performance of neural networks on some image classification
  tasks.%2 background
  These results raise the question of whether neural networks only learn
  successfully if kernels also learn successfully, despite neural nets being more expressive. %3 main question
  Here, we show theoretically that two-layer neural networks (2LNN) with
  only a few neurons can beat the performance of kernel learning on a simple Gaussian mixture classification task. %4 main result
  We study the high-dimensional limit,
  i.e.~when the number of samples is linearly proportional to the dimension, and
  show that while small 2LNN achieve near-optimal performance on this task, lazy
  training approaches such as random features and kernel methods do
  not.% 5 Detailed results 
  Our analysis is based on the derivation of a closed set of equations that track the learning dynamics of the 2LNN and thus allow
  to extract the asymptotic performance of the network as a function of
  signal-to-noise ratio and other hyperparameters. %6 Detailed results 2
  We finally illustrate how over-parametrising the neural network leads to faster
  convergence, but does not improve its final performance. % 7 Det. results 3
\end{abstract}

%\balance
\section{Introduction}
Explaining the success of deep neural networks in many areas of machine learning
remains a key challenge for learning theory. A series of recent theoretical
works made progress towards this goal by proving trainability of two-layer
neural networks (2LNN) with gradient-based methods~\cite{jacot2018neural,
  allen2018convergence, Li2018a, allen2019convergence, cao2019generalization,
  du2019gradient}. These results are based on the observation that strongly
over-parameterised 2LNN can achieve good performance even if their first-layer
weights remain almost constant throughout training. This is the case if the
initial weights are chosen with a particular scaling, which was dubbed the
``lazy regime'' by~\textcite{chizat2019lazy}. Going a step further, simply
fixing the first-layer weights of a 2LNN at their initial values yields the
well-known random features model of~\textcite{rahimi2008random,
  rahimi2009weighted}, and can be seen as an approximation of kernel
learning~\cite{scholkopf2018learning}. This behaviour is to be contrasted with
the ``feature learning regime'', where the weights of the first layer move
significantly during training. Recent empirical studies showed that on some
benchmark data sets in computer vision, kernels derived from neural networks
achieve comparable performance to neural networks~\cite{matthews2018gaussian,
  lee2018deep, garriga-alonso2019deep, arora2019exact, li2019enhanced,
  shankar2020neural}.
  
These results raise the question of whether neural networks only learn
successfully if random features can also learn successfully, and have led to a
renewed interest in the exact conditions under which neural networks trained
with gradient descent achieve a better performance than random
features~\cite{bach2017breaking,
  yehudai2019power, wei2019regularization, li2020learning, daniely2020learning,
  geiger2020disentangling, paccolat2021geometric,
  suzuki2020benefit}. \textcite{chizat2020implicit} studied the implicit bias of
wide two-layer networks trained on data with a low-dimensional structure. They
derived strong generalisation bounds, which, when both layers of the network are
trained, are independent of ambient dimensions, indicating that the network is
able to adapt to the low dimensional structure. In contrast, when only the
output layer of the network is trained, the network does not possess such an
adaptivity, leading to worse performance. \textcite{ghorbani2019limitations,
  ghorbani2020neural} analysed in detail how data structure breaks the curse of
dimensionality in wide two-layer neural networks, but not in learning with
random features, leading to better performance of the former.

\begin{figure*}[t!]
  \centering
  \includegraphics[width=\linewidth]{{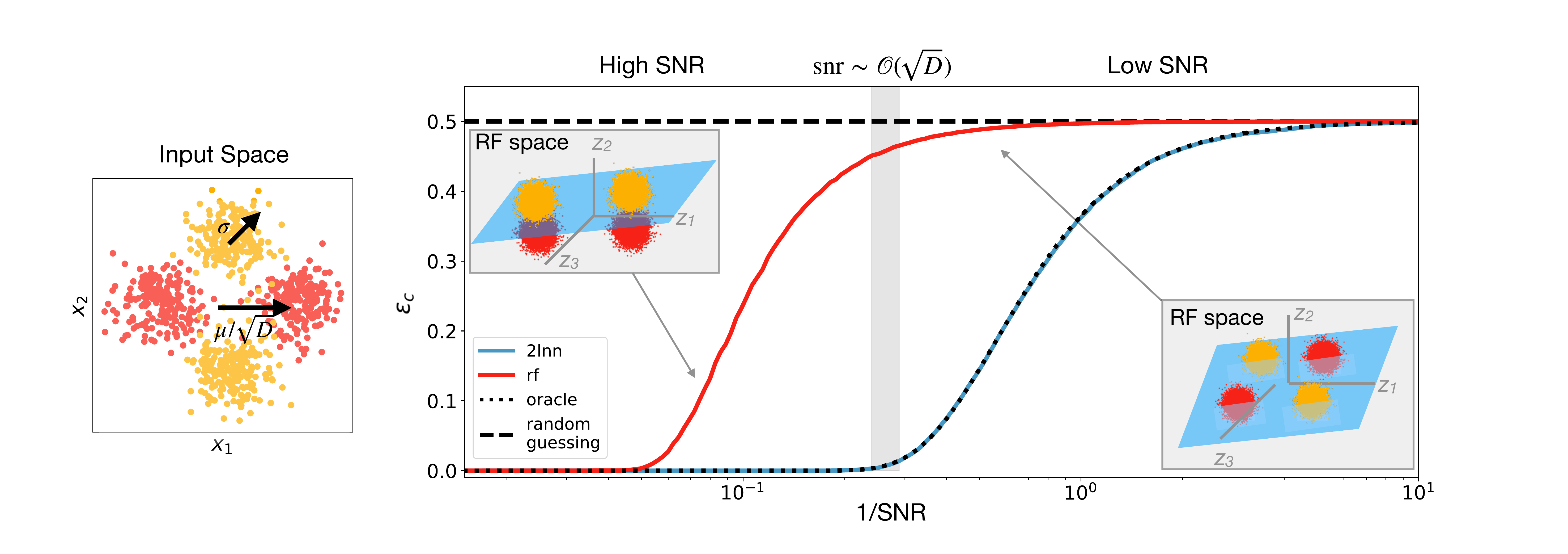}}
\vspace{-0.5cm}
  \caption{\label{fig:snr_error} \textbf{Random Features and 2LNN on
      high-dimensional Gaussian mixture classification} \emph{\textbf{(Left)}}
    Consider a data distribution that is a mixture of four Gaussians in $D$
    dimensions. The first two components of the centroids organized in a
    XOR-like manner as shown, while the other $D-2$ directions of the centroids
    are set to zero. The signal-to-noise is
    $\snr = \nicefrac{|\mu|}{\sqD \sigma}$. We used $\mu = \sqrt{D}$ so the
    $\snr$ is effectively given by the inverse of the width of the
    Gaussian: $\snr = 1/\sigma$.  \emph{\textbf{(Right)}} A two-layer neural
    network with $K=4$ hidden neurons and ReLU non-linearity trained using
    stochastic gradient descent, achieves a long-time test error close to the
    optimal (oracle) error~\eqref{eq:ec_oracle} for the whole range of $\snr$. The
    test error is obtained analytically using techniques described in
    Sec.~\eqref{sec:state-evolution}. In sharp contrast, random features (RF),
    whose performance is given by
    Eq.~\eqref{eq:rf_asymptotic_classification_error}, require a high $\snr$ to
    perform as well as the oracle. They performs better than chance when
    $\snr \gg \sqrt{D}/\min(P,N)^{1/4}$, and thus requires a diverging
    $\snr$ in the high-dimensional limit. The insets show the mixture
    \emph{after} applying random features; only at high $\snr$ does the mixture
    become linearly separable. \emph{Parameters:} RF error is computed with
    $D=10000$ and $P=2D$ and the 2LNN's with $D=1000$. For both methods
    $\eta=0.1$, $\nicefrac{|\mu|}{\sqD}=1$.  }
\end{figure*}
%\vspace{-0.8cm}

\subsection{Main contributions} We show that even a two-layer neural
network with \emph{only a few hidden neurons} outperforms kernel methods on the
classical problem of Gaussian mixture classification. We give a sharp asymptotic
analysis of 2LNN and random features on Gaussian mixture classification in the
high-dimensional regime where the number of samples $N$ is linearly proportional
to the input dimension $D\to\infty$. More precisely:

\begin{enumerate}[label=(\roman*),wide = 0pt] % wide = 0pt
\item\label{contrib:ode} We analyse 2LNN by deriving a closed set of ordinary
  differential equations (ODEs), which track the test error of 2LNN with a few
  hidden neurons $K\!\sim\! O(1)$ trained using one-pass (or online) SGD on
  Gaussian mixture classification. We thereby extend the classical ODE analysis
  of~\textcite{Riegler1995, Saad1995a} where the label $y(x)$ is a function of
  the input $x$, to a setup where the input is conditional on the label. Solving
  these equations for their asymptotic fixed point, i.e.~taking $t\to \infty$
  \emph{after} $D\to \infty$, yields the final classification error of the 2LNN.%,
  % such as the blue line in Fig.~\ref{fig:snr_error}.
\item Keeping in mind the high-dimensional limit  where the
  number of samples $N$ is proportional to~$D$, 
  We analyse how Gaussian mixtures
  are transformed under $P$ random features in the regime $P,D\to\infty$ with
  $\gamma \equiv \nicefrac{P}{D}$ fixed. In the high-dimensional limit the
  performance at large $\gamma$ converges to the one of the corresponding
  kernel~\cite{rahimi2008random, rahimi2009weighted, elkaroui2010spectrum,
    pennington2017nonlinear, louart2018random, liao2018spectrum}, and we can thus recover the performance of kernel learning by taking $\gamma$ large enough.%, which as argued in \cite{mei2021generalization}, occurs when $P$ is roughly equal to $N$.
\item\label{contrib:RF} We compute the asymptotic generalisation  of random
  features on mixtures of Gaussians, which allows us to compare their performance to the performance of
  2LNN for various signal-to-noise ratios.
\end{enumerate}
While we do not attempt formal rigorous derivations, to keep the paper readable, our theoretical claims are however amenable to rigorous theorems. In particular the ODEs analysis could be formalised rigorously using the technique of~\cite{wang2019solvable, goldt2019dynamics}. Our results are valid for generic Gaussian Mixtures with $O(1)$ clusters and we focus on the particular example of the XOR-like mixture in order to make the problematic clear.

\subsection{A paradigmatic example}
Our results can be illustrated with a data distribution where inputs are
distributed in a mixture of four Gaussians, whose centres form a XOR-function,
and thus cannot be linearly separated in direct space. We find that neural
networks with a few neurons have no problems learning a good partitioning of the
space in this situation, reaching oracle-like performance in the process. Kernel
methods, however, manage to do so only if the centres are extremely well
separated, and completely fail when they are too close. This is illustrated in
Fig.~\ref{fig:snr_error}: Inputs $x=(x_r)\in\reals^D$ are drawn from the
Gaussian mixture shown on the left, where inputs in red, yellow have labels
$y=~1,-1$, respectively. The first two components of the means are organised
as in the diagram, while the remaining $D-2$ components are zero, yielding a
XOR-like pattern. Each Gaussian cloud has standard deviation~$\sigma \mathbb{I}_D$, as Gaussian noise is added to all components of the input,
resulting in an $\snr$ of $\nicefrac{|\mu|}{\sqrt{D \sigma^2}}$.

We compare the
performance of a \textbf{two-layer neural network}~$\phi_\theta$ with parameters
$\theta= (K, v, W, g)$,
\begin{align}
 \label{eq:2lnn}
  \phi_{\theta}(x )  = \sum_{k=1}^K  v^k\,  g \left( \lambda^k\right) , \quad
  \lambda^k \equiv \ussD \sum_{r=1}^D w^k_r x_r,
\end{align}
where $v = (v^k) \in \reals^K$ and $W = (w^k_r) \in \reals^{K \times D}$ are the
weights of the network and $g: \reals \to \reals$ is a non-linear function
shown above only has $K=4$ neurons, and we keep $K$ of order 1 compared to the
input dimension~$D~\to~\infty$ throughout this paper. We train the 2LNN using
online stochastic gradient descent, where at each step of the algorithm, we draw
a new sample from the mixture. We study the high-dimensional limit 
$
  t \equiv \nicefrac{N}{D} = O(1),$
and obtain the final performance of the 2LNN in the limit $t\to \infty$
(after $D\to \infty$) of the ODEs we derive in Sec.~\ref{sec:odes}. In blue, we
plot the final classification error
\begin{equation}
  \label{eq:ec}
  \eclass{ }(\theta) = \mathop{\EE}%_{q(x, y)}
  \Theta \left[ -y \phi_{\theta}(x) \right],
\end{equation}
where the expectation $\EE$ is computed over the Gaussian mixture for a network
with fixed parameters~$\theta$ and $\Theta$ is the Heaviside step function. The
classification error of the 2LNN is very close to that of an \emph{oracle} with
knowledge of the means of the mixture that assigns to each input the label of
the nearest mean, achieving a classification error of
\begin{equation}
  \label{eq:ec_oracle}
  %\epsilon_c^{\text{oracle}} =1-
  %\frac{1}{2}\left(1+\operatorname{erf}\left(\frac{1}{2\sigma}\right)^2\right),
  \epsilon_c^{\text{oracle}} = 
  \nicefrac{1}{2}\left(1-\operatorname{erf}\left(\nicefrac{|\mu|}{2\sigma\sqD}\right)^2\right).
\end{equation}

We compare the performance of the 2LNN to the performance of \textbf{random
  features}~\cite{rahimi2008random, rahimi2009weighted}, where we first project
the inputs $x$ to a higher-dimensional feature space, where features
$z=(z_i)\in \reals^P$ are given by
\begin{equation}
    \label{eq:rf_z}
    z_i=\psi(u_i), \qquad u_i \equiv \sum_{r=1}^D\ussD F_{ir}x_r\, ,
\end{equation}
where $F\in \reals^{P \times D}$ is a random, but fixed projection matrix and
$\psi: \mathbb R\to \mathbb R$ is an element-wise non-linearity. The features
are then fit by training a linear model,
\begin{align}
  \label{eq:rf_phi}
  \phi_\theta(z)= \frac{1}{\sqrt{P}} \sum_{i=1}^{P}  w_i z_i,
\end{align}
with weights $w\in\mathbb{R}^{P}$, using SGD, where we again draw a fresh sample from the mixture to evaluate the
gradients at each step. The performance of RF is shown in red in
Fig.~\ref{fig:snr_error}. While RF achieve low classification error at high $\snr$,
there is a wide range of $\snr$ where random features do significantly
worse than the 2LNN. The insets give the intuition behind this result: at high
$\snr$, RF map the inputs into  linearly separable mixture in random
feature space (left) while at lower ${\rm snr}$, the transformed mixture is not linearly separable in RF space (right), leading to poor performance.  %The transition between the two regime occurs when

We emphasise that we study random features in
the \emph{high-dimensional limit} where we let $N,D\to \infty$ with their
ratio~$t=\nicefrac{N}{D}\!\sim\!O(1)$ as before, while also letting the number
of random features $P \to \infty$ with their ratio
$\gamma\equiv\nicefrac{P}{D} \sim O(1)$
fixed. This regime has been studied in a series of recent
works~\cite{lelarge2019asymptotic, couillet2019high, liao2019innerproduct,
  mai2019high, deng2019model, kini2020analytic, mignacco2020role}. While we
concentrate on random features, we note that we can recover the performance of
kernel methods~\cite{rahimi2008random,rahimi2009weighted} by sending~$\gamma \to \infty$. Indeed, as $\gamma = \nicefrac{P}{D}$ grows, the gram
matrix converges to the limiting kernel gram matrix in the high-dimensional
regime; detailed studies of the convergence in this regime can be found in~\cite{elkaroui2010spectrum, pennington2017nonlinear, louart2018random,
  liao2018spectrum}. We can thus recover the performance for any general
distance or angle based kernel method, e.g.~the NTK
of~\textcite{jacot2018neural}, by considering $\gamma$ large enough in our
computations with random features. Note, however, that this must be done with
some care. Our results for random projections are given for $N > P$. As
discussed by~\textcite{ghorbani2019limitations, mei2021generalization}, the
relevant dimension for random features performances is, rather than $P$, the
minimum between $N$ and $P$.
%% I am not sure what is meant in the next two sentences
Since we focus here in the regime where %$N$ is $O(D)$,
increasing $P$ beyond $O(D)$, and therefore $\gamma$ beyond $O(1)$, is not allowed. Indeed, we shall see that Lazy training methods such as kernels or random projections require asymptotically $N=O(D^2)$ samples to beat a random guess, while neural-networks achieves oracle-like performances with only~$N=O(D)$ samples.

\paragraph{Reproducibility} \label{sec:reproducibility} We provide code to reproduce our plots and 
solve the equations of Sec.~\ref{sec:odes} 
at~
%\url{https://github.com/mariaref/rfvs2lnn_GMM_online}[github.com/mariaref/rfvs2lnn_GMM_online].
 \href{https://github.com/mariaref/rfvs2lnn_GMM_online}{github.com/mariaref/rfvs2lnn{\_}GMM{\_}online}.

\subsection{Further related work}
\label{sec:further-work}

\paragraph{Separation between kernels \& 2LNN} \textcite{barron1993universal}
already discussed the limitations of approximating functions with a bounded
number of random features within a worst-case
analysis. \textcite{yehudai2019power} construct a data distribution that can be
efficiently learnt by a single ReLU neuron, but not by random features.
\textcite{wei2019regularization} studied the separation between 2LNN \& RF and show the \emph{existence} of a small
($K\sim O(1)$) network that beats kernels on this data distribution, and study
the \emph{dynamics} of learning in the same mean-field limit as
\textcite{chizat2020implicit} and~\textcite{ghorbani2019limitations,
  ghorbani2020neural}. Likewise, \textcite{li2020learning} show separation
between kernels \& neural networks in the mean-field limit on the phase
retrieval problem. \textcite{geiger2020disentangling}
 investigated numerically the role of architecture and data 
in determining whether lazy or feature learning perform
better. \textcite{paccolat2021geometric} studied how neural networks can
compress inputs of effectively low-dimensional data.

\paragraph{Gaussian mixture classification} is a well-studied problem in
statistical learning theory, and its supervised version was recently considered
in a series of works from the perspective of Bayes-optimal
inference~\cite{lelarge2019asymptotic, mai2019high,
  deng2019model}. \textcite{mignacco2020role, mignacco2020dynamical} studied the
dynamics of stochastic gradient descent on a finite training set using dynamical
mean-field theory for the perceptron, which corresponds to the case~$K=1,v^1=1$
in Eq.~\eqref{eq:2lnn}.  \textcite{liao2019innerproduct} and
\textcite{couillet2019high} studied  mixture classification with kernel in an unsupervised setting using random matrix theory.

\paragraph{Dynamics of 2LNN} A classic series of papers by \textcite{Biehl1995}
and~\textcite{Saad1995a} studied the dynamics of 2LNN as in Eq.~\eqref{eq:2lnn}
trained using online SGD in the classic teacher-student
setup~\cite{Gardner1989}, where inputs $x$ are element-wise i.i.d.~Gaussian
variables and labels are obtained from a ``teacher'' network with random
weights. They derived a set of closed ODEs that track the test error of the
student (see also~\textcite{Saad1995b, Biehl1996, saad2009line} for further
results and~\textcite{goldt2019dynamics} for a recent proof of these equations).
There have been several extensions of this approach to different data
distributions~\cite{yoshida2019datadependence, goldt2020modelling,
  goldt2020gaussian}. All of these works, though, consider the label~$y$ as a
function of the input $x$, or as a function of a latent variable from which $x$
is generated. Here, we extend this type of analysis to a case where the input is
conditional on the label, a point of view taken implicitly
by~\textcite{cohen2020separability}.

The reduction of the dynamics to a set of low-dimensional ODEs should be
contrasted with the ``mean-field'' approach, where the number of hidden neurons
$K$ is sent to infinity while the input dimension $D$ is kept finite. In this
limit, the neural networks are still a more expressive function class than the
corresponding reproducing kernel Hilbert space~\cite{Chizat2018, Sirignano2018,
  Rotskoff2018, Mei2018}.  The evolution of the network parameters in this limit
can be described by a high-dimensional partial differential equation. This
analysis was used in the aforementioned works by
\textcite{ghorbani2019limitations, ghorbani2020neural}.
\section{Neural networks for GM classification}

\subsection{Setup}

We draw inputs $x\!=\!(x_i)\!\in\!\reals^D$ from a high-dimensional Gaussian
mixture, where all samples from one Gaussian are assigned to one of two possible
labels $y=\!\pm 1$, which are equiprobable. The data distribution is thus
\begin{equation}
  \label{eq:data}
  q(x, y)\! = \!q(y) q(x | y), \quad \!\!\! q(x | y) \!=\!\!\!\!\!
  \sum_{\alpha\in\mathcal{S}\left(y\right)}\!\!\!\!\! \mathcal{P}_\alpha \normal_\alpha(x)\, ,% \left(\frac{\gamma\mu^\alpha}{\sqrt{D}}, \Omega^\alpha\right)
\end{equation}
where $\normal_\alpha(x)$ is a multivariate normal distribution with mean
$\nicefrac{\mu^\alpha}{\sqrt{D}}$ and covariance $\Omega^\alpha$.  The index set
$\mathcal{S}(y)$ contains all the Gaussians that are associated with the
label~$y $. We choose the constants $\mathcal{P}_\alpha$ such that
$q(x, y)$ is correctly normalised. To simplify notation, we focus on binary
classification, which can be learnt using a student with a single output
unit. Extending our results to $C$-class classification, where the student has
$C$ output heads, is
straightforward. % We note that the dynamics of two-layer neural networks with multiple output units trained in the classical teacher-student setup were recently analysed in the same limit by~\textcite{yoshida2019statistical}.

\paragraph{Training} The network is trained using stochastic gradient descent on
the \emph{quadratic error} for technical reasons related to the analysis. The update equations for the weights at the $\mu$th step of the algorithm,
$\dd w^k_i \equiv \left(w^k_i\right)_{\mu+1}- \left(w^k_i\right)_{\mu}$, read
\begin{subequations}
  \label{eq:sgd}
  \begin{align}
    \dd w^k_i & =-\frac{\eta}{\sqrt{D}} v^k \Delta g'(\lambda^k) x_i - \frac{\eta}{\sqrt{D}}\kappa w^k_i , \label{eq:sgdw}\\
    \dd v^k & = - \frac{\eta}{D} g(\lambda^k) \Delta- \frac{ \eta}{D}\kappa v^k\, , \label{eq:sgdv}
  \end{align}
\end{subequations}
where $\Delta \!=\! \sum_{j = 1}^K\! v^j g(\lambda^j)\! -\!y$ and
$\kappa\! \in \!\mathbb{R}$ is a $L_2$-regularisation constant. Initial weights are taken i.i.d.~from the normal distribution with standard deviation $\sigma_0$. The
different scaling of the learning rates $\eta$ for first and second-layer
weights guarantees the existence of a well-defined limit of the SGD
dynamics as $D\!\to\!\infty$. We make the crucial assumption that at each step
of the algorithm, we use a previously unseen sample $(x, y)$ to compute the
updates in Eq.~\eqref{eq:sgd}. This limit of
infinite training data is variously known as online learning or
one-shot/single-pass SGD.

\subsection{Theory for the learning dynamics of 2LNN}
\label{sec:odes}

\paragraph{Statics} Since we are training on the quadratic error, the first step
of our analysis is to rewrite the prediction mean-squared error $\pmse$ as a sum
over the error made on inputs from each Gaussian $\alpha$ in the mixture,
\begin{align}
\begin{split}
  \pmse(\theta) & = \mathop{\EE}_{q(x, y)} \left(y - \phi_{\theta}(x)
                  \right)^2 \label{eq:pmse}\\
  & = \sum_{y} \sum_{\alpha\in\mathcal{S}\left(y_i\right)} \!\!\! q(y_i)
  \mathcal{P}_\alpha \mathop{\EE}_{\alpha} {\left[ \sum_{k} v^k
      g(\lambda^k) - y\right]}^2\,
\end{split}
\end{align}
where the average $\displaystyle\mathop{\EE}_{\alpha}$ is taken over the
$\alpha$th normal distribution $\mathcal{N}_\alpha$ for fixed parameters
$\theta$. To evaluate the average, notice that the input~$x$ only enters the
expression via products with the student weights~$\lambda\!=\!(\lambda^k)$; we can
hence replace the high-dimensional averages over $x$ with an average over the
$K$ ``local fields''~$\lambda^k$. An important simplification occurs since the
$\lambda^k$ are jointly Gaussian when averages are evaluated over just a single
distribution in the mixture. We write the first two moments of the local fields
as $M\!=\!(M^{k}_\alpha)$ and~$Q\!=\!(Q^{k\ell}_\alpha)$, with
\begin{subequations}
  \label{eq:OP}
  \begin{align}
    M^{k}_\alpha&\equiv\mathop{\EE}_{\alpha} \lambda^k= \frac{1}{D} \sum_{r}w^k_r
                  \mu^\alpha_r,\label{eq:M} \\
    Q^{k\ell}_\alpha&\equiv\mathop{\cov}_{\alpha}
                      \left(\lambda^k, \lambda^\ell\right)=\frac{1}{D}\sum_{r, s}w^k_r
                      \Omega^\alpha_{rs} w^\ell_s \label{eq:Q} .
  \end{align}
\end{subequations}
Any average over a Gaussian distribution is a function of only the first two
moments of that distribution, so the $\pmse$ can be written as a function of the
``order parameters''~$M$ and $Q$ and of the $K\!\sim\! O(1)$ second-layer
weights $v\!=\!(v^k)$:
\begin{equation}
  \lim_{D\to\infty} \pmse(\theta) \to \pmse(Q, M, v).
\end{equation}
Likewise, the classification error $\epsilon_c$~\eqref{eq:ec} can also be
written as a function of the order parameters only:
$\mathop{\lim}_{D\to\infty} \epsilon_c(\theta)\! \to\! \epsilon_c(Q, M, v)$. The order
parameters have a clear interpretation:~$M^{k}_\alpha$~encodes the overlap
between the $k$th student node and the mean of the $\alpha$ cluster, and plays a
similar role to the teacher-student overlap in the vanilla teacher-student
scenario.  $Q^{kl}_\alpha$ instead tracks the overlap between the various
student weight vectors, with the input-input covariance $\Omega^\alpha$
intervening. The strategy for our analysis is thus to derive equations that
describe how the order parameters $(Q, M, v)$ evolve during training, which will
in turn allow us to compute the $\pmse$ of the network at all times.

\begin{figure*}[t!]
  \centering
%  \vspace{-3mm}
  \includegraphics[width=.7\linewidth]{{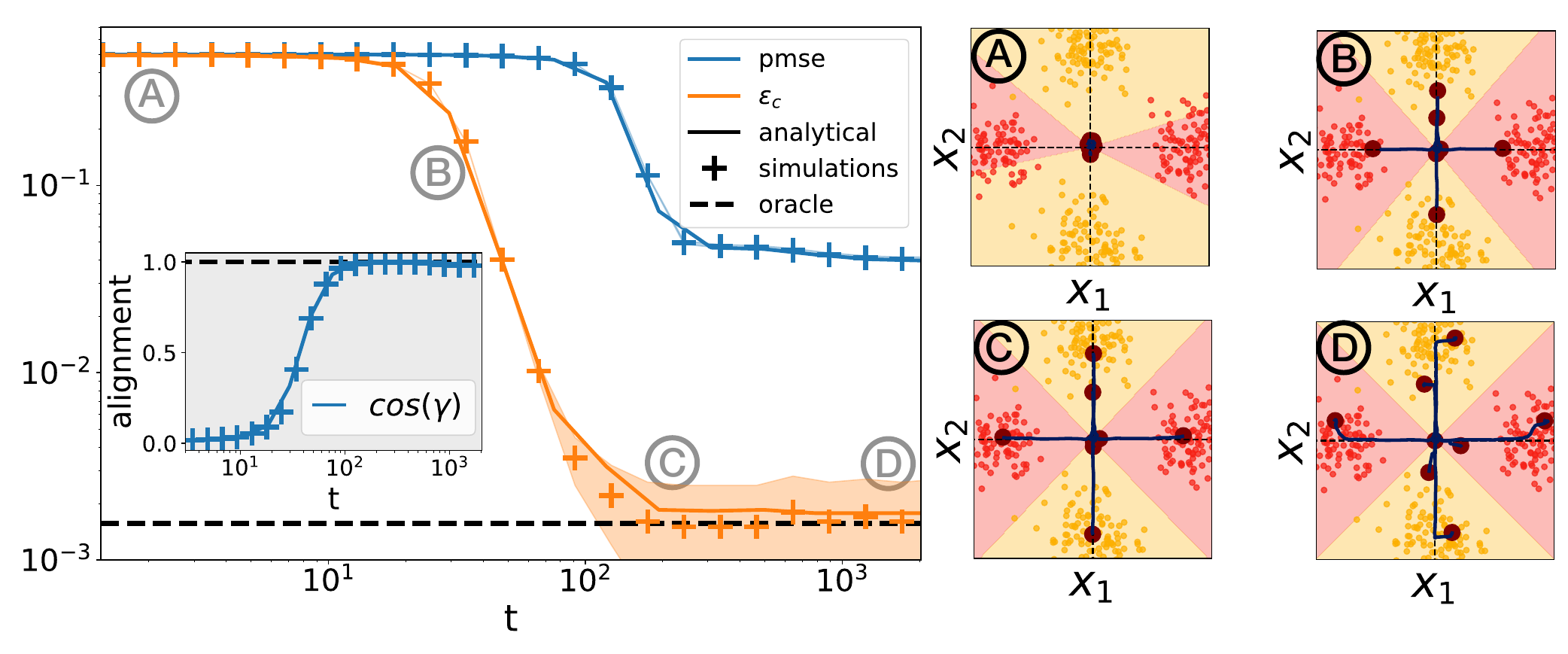}}
  \vspace{-3mm}
  \caption{\label{fig:weights_evolution} \textbf{How the 2LNN learns a XOR-like
      GM} \textbf{(Left)} Evolution of the prediction mean-squared error
    $\pmse$~\eqref{eq:pmse} and the classification error~\eqref{eq:ec} of a 2LNN
    with $K=8$ neurons trained on the XOR-like mixture of
    Fig.~\ref{fig:snr_error}. We plot the test errors as obtained from a
    single simulation with $D=1000$ (crosses) and from integration of the ODEs
    of Sec.~\ref{sec:odes}. The dashed black line is the classification error of
    an oracle with knowledge of the means $\mu^\alpha$,
    Eq.~\eqref{eq:ec_oracle}. The inset shows the mean angle of the network
    weights to the means of the mixture. \textbf{(Right)} Projections of the
    first layer weights (dots) onto the plane spanned by the means of the XOR-like
    mixture at different times during training. Shaded areas indicate the
    decision boundaries of the network, where its output $\phi_\theta(x)$
    changes sign.  \emph{Parameters:} $K=8, D=1000, \sigma=0.05, \eta=0.1$
    weights initialised with s.t.d. $\sigma_0=1$, $\kappa=10^{-2}$.}
     \vspace{-3mm}
\end{figure*}

\paragraph{Dynamics} We derived a closed set of ordinary differential equations
that describe the evolution of the order parameters in the case where each
Gaussian in the mixture has the same covariance matrix $\Omega$.
We proceed here with a brief statement of the equations and deffer the detailed derivation to Sec.~\ref{app:derivODEs}.
The approach is most easily illustrated with
the second-layer weights $v^k$. The key idea to compute the average change in
the weight $v^k$ upon an SGD update~\eqref{eq:sgdv}, $\dd v^k$, which can be decomposed into a contribution from every Gaussian in the mixture,
% SG: please keep this as finite differences at this stage
\begin{equation}
  \EE \dd v^k\! =\!\sum_{\alpha\in\mathcal{S}\left(+\right)} \mathcal{P}_\alpha
  \dd v^k_{\alpha^+} +\sum_{\alpha\in\mathcal{S}\left(-\right)}
  \mathcal{P}_\alpha \dd v^k_{\alpha^-}\, ,
\end{equation}
where the change $\dd v^k_{\alpha^+}$ is obtained directly from Eq.~\eqref{eq:sgdv},
\begin{equation}
  \dd v^k_{\alpha}\!=\!\frac{\eta}{D} \mathop{\EE}_{\alpha} y_\alpha g(\lambda^k)\! -\! \frac{\eta}{D} \sum_j\! v^j \mathop{\EE}_{\alpha} g(\lambda^k)g(\lambda^j)\!-\!\frac{\eta}{D}\kappa v^k.
\end{equation}
The averages that remain to be computed only involve the true label and the local fields $\lambda$. The former is a constant within each Gaussian while the latter are jointly Gaussian. It follows, that also these averages can be
expressed in terms of only the order parameters and the equation
closes. As we discuss in the appendix, in the high-dimensional
limit $D\!\to\!\infty$ the normalised
number of samples $t\!\equiv\! \nicefrac{N}{D}$ can be interpreted as a continuous
time, which allows the dynamics of $v^k$ to be captured by the ODE~\eqref{eq:eom-v}.

The order parameters $Q$ require an additional step which consists in
 diagonalising the sum  $Q^{k\ell}~\!\sim~\!\mathop{\sum}_{r, s}^D w^k_r \Omega_{rs} w^\ell_s$ by
introducing the integral representation
\begin{equation}
  \label{eq:Q_int}
  Q^{kl}=\int \dd \rho \;  p_{\Omega}(\rho)  \;\rho q^{kl}(\rho),\quad
\end{equation}
where $p_{\Omega}(\rho)$ is the spectral density of $\Omega$, and $q^{kl}(\rho)$
is a density whose time evolution can be characterised in the thermodynamic
limit. We relegate the full expression of the equation of motion for
$q^{kl}(\rho)$ to Eq.~\eqref{eq:eom-q} of the appendix. Crucially, it involves
only averages that can be expressed in terms of the order
parameters~\eqref{eq:OP}, and hence the equation closes. Likewise, the order
parameter $M$ can be rewritten in terms of a density as
%\begin{equation}
$  M^{\alpha k} =\int \dd \rho \; p_{\Omega}(\rho)  m^{\alpha k}(\rho)$.
%\end{equation}
The dynamics of $m^{\alpha k}$ is described by Eq.~\eqref{eq:eom-m}.

\paragraph{Solving the equations of motion} The equations are valid for any mean
and covariance matrix~$\Omega$. Solving them requires evaluating
multidimensional integrals of dimension up to 4, e.g.~$\mathop{\EE}_{\alpha} g'(\lambda^k) \lambda^\ell y$, which can be effeciently
estimated using Monte-Carlo (MC) methods. We provide a ready-to-use numerical
implementation on the GitHub.

\paragraph{Comparing theory and simulation} On the left of
Fig.~\ref{fig:weights_evolution}, we plot the evolution of the
$\pmse$~\eqref{eq:pmse} and the classification error~\eqref{eq:ec} of a 2LNN
with $K=8$ neurons trained on the XOR-like mixture of
Fig.~\ref{fig:snr_error}. We plot the test errors obtained from integration of
the order parameters with solid lines, and the same quantities computed using a
test set during the simulation with crosses. The agreement between ODE
predictions and a single run of SGD is good, even at intermediate system size
($D=1000$). In the App.~\ref{app:derivODEs}, we give additional plots for the
simulated dynamics of the individual order parameters and find very good
agreement with predictions obtained from the ODEs
(cf.~Fig.~\ref{fig:ode-vs-sim}). Note that although we initialise the weights of
the student randomly and independently of the means, there is an initial overlap
between student weights and the means of order $1/\sqrt{D}$ due to finite-size
fluctuations. To capture this with the ODEs, we initialise them in a regime of
weak recovery, where~$M^k_\alpha\!\neq\!0$. For a detailed discussion of the early
period of learning up to weak recovery, see~\textcite{arous2020classification}.

\paragraph{How 2LNNs learn the XOR-like mixture} A closer look at the learning
dynamics on the right of Fig.~\ref{fig:weights_evolution} reveals several phases
of learning. There we show the first-layer weight vectors of the 2LNN, projected
into the plane spanned by the four means of the mixture, at four different times during
training. The regions shaded in red and yellow indicate the decision boundaries
of the network, which correspond to the line where the network's output
$\phi_\theta(x)$ changes its sign. A 2LNN with $K\ge 4$ neurons can approach the
classification error of the oracle~\eqref{eq:ec_oracle} if its weight vectors approach the four means, with corresponding second-layer weights. Panel (C) shows that network reaches this configuration. However, this configuration does not minimise the mean-squared error used during training~\eqref{eq:sgd}, so eventually the weights depart slightly from the
means to converge to a solution with lower mean squared error (D). This is confirmed by the inset on the left of Fig.~\ref{fig:weights_evolution}, where
we see that the average angle of the network weights to the means has a maximum around $t=300$, before decaying slightly at the end of training.

\subsection{Predicting the long-time performance of 2LNN}
\label{sec:state-evolution}

\begin{figure}[t!]
\centering
  \includegraphics[width=7.5cm]{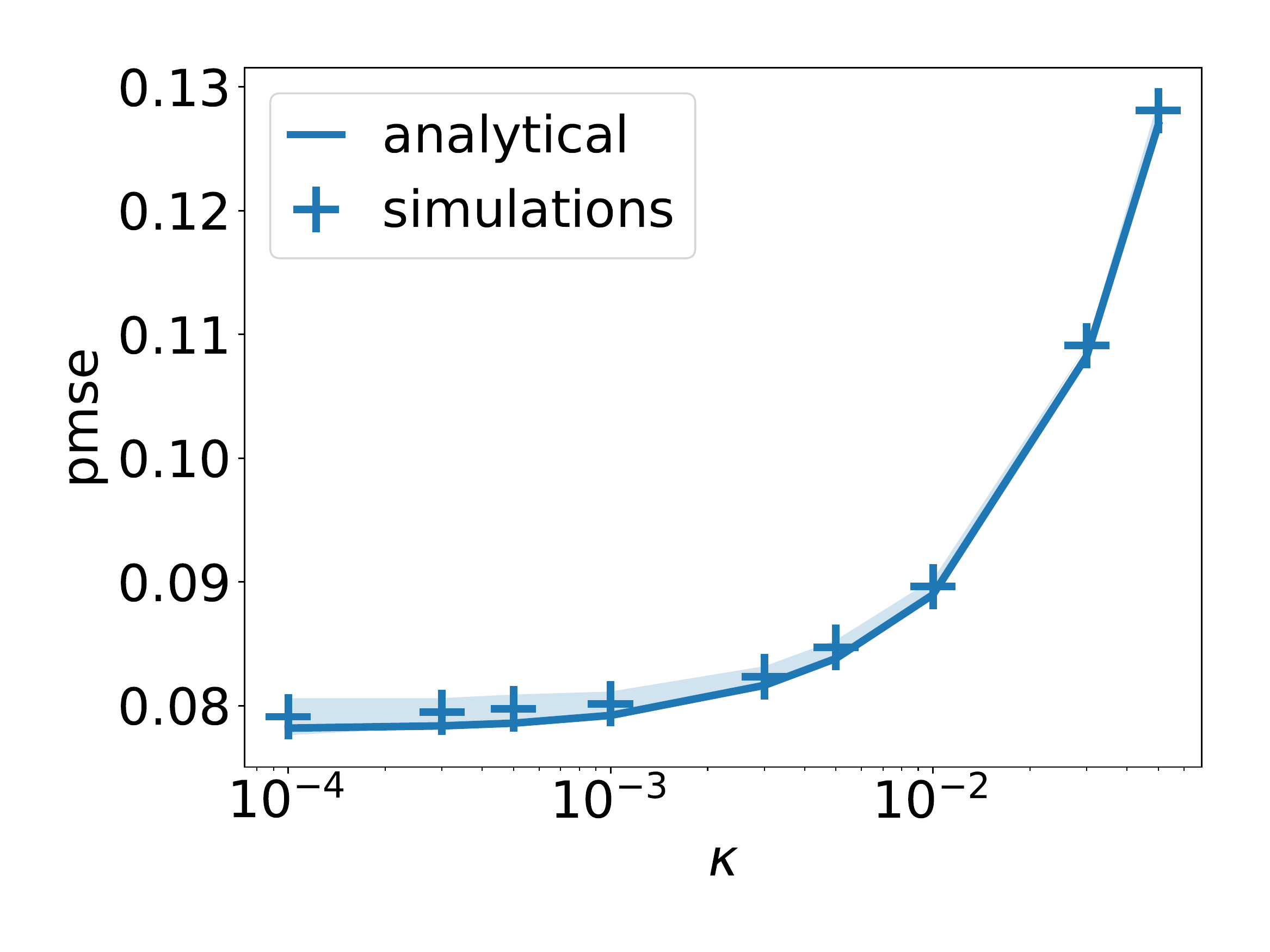}
%  \vspace{-5mm}
\vspace{-0.5cm}
  \caption{\label{fig:SE_var_reg} Prediction mean-squared error~\eqref{eq:pmse}
    on the XOR-like mixture versus weight decay. Results obtained from
    a fixed point analysis of the ODEs with 4 degrees of freedom discussed in
    Sec.~\ref{sec:state-evolution} ($\sigma^2=0.1, K=4, \eta=0.1, 10^4$) Monte-Carlo samples. The shaded area indicates standard deviation over 10 runs. }
    \vspace{-0.5cm}
\end{figure}

Direct integration of the ODEs is numerically expensive. A more straight forward way
to extract information from the ODEs is to find their asymptotic fixed point,
which fully characterises the $t \to \infty $ performance of the network. However, the
number of equations is already 26 for a 2LNN
with 4 neurons trained on the XOR mixture, and scales like $K^2$. The key to finding fixed
points efficiently is thus to make an \emph{ansatz} with fewer degrees of
freedom for the matrices $Q$ and $M$ which solve the equations.  For example,
one could impose $Q^{kk}=Q$ and $Q^{k\ell}=C, k\neq \ell$.  By exploiting the symmetries of the XOR-like mixture, we
find that the fix points of the equations can be described by only $K = 4$
parameters: $\nicefrac{K}{2}$ angles between weight vectors and means, and
$\nicefrac{K}{2}$ norms, as described in
Appendix~\ref{app:ansatz-for-XOR}. Finding the fixed-point of this reduced
four-dimensional system allows to compute, for example, the dependence of the
generalisation error as a function the regularisation in
Fig.~\ref{fig:SE_var_reg}. The agreement between simulation and analytical
predictions is again good, and we find that increasing the regularisation only
increases the test error of the student. This is in agreement with previous work
on two-layer networks in the same limit in the  teacher-student
setup, where $L_2$-regularisation was also found to hurt
performance~\cite{Saad1997}. 
%We provide an implementation of the fixed point finder at \url{https://github.com/mariaref/rfvs2lnn_GMM_online}.

\begin{figure}[h]
  \centering
  \includegraphics[width=7.5cm]{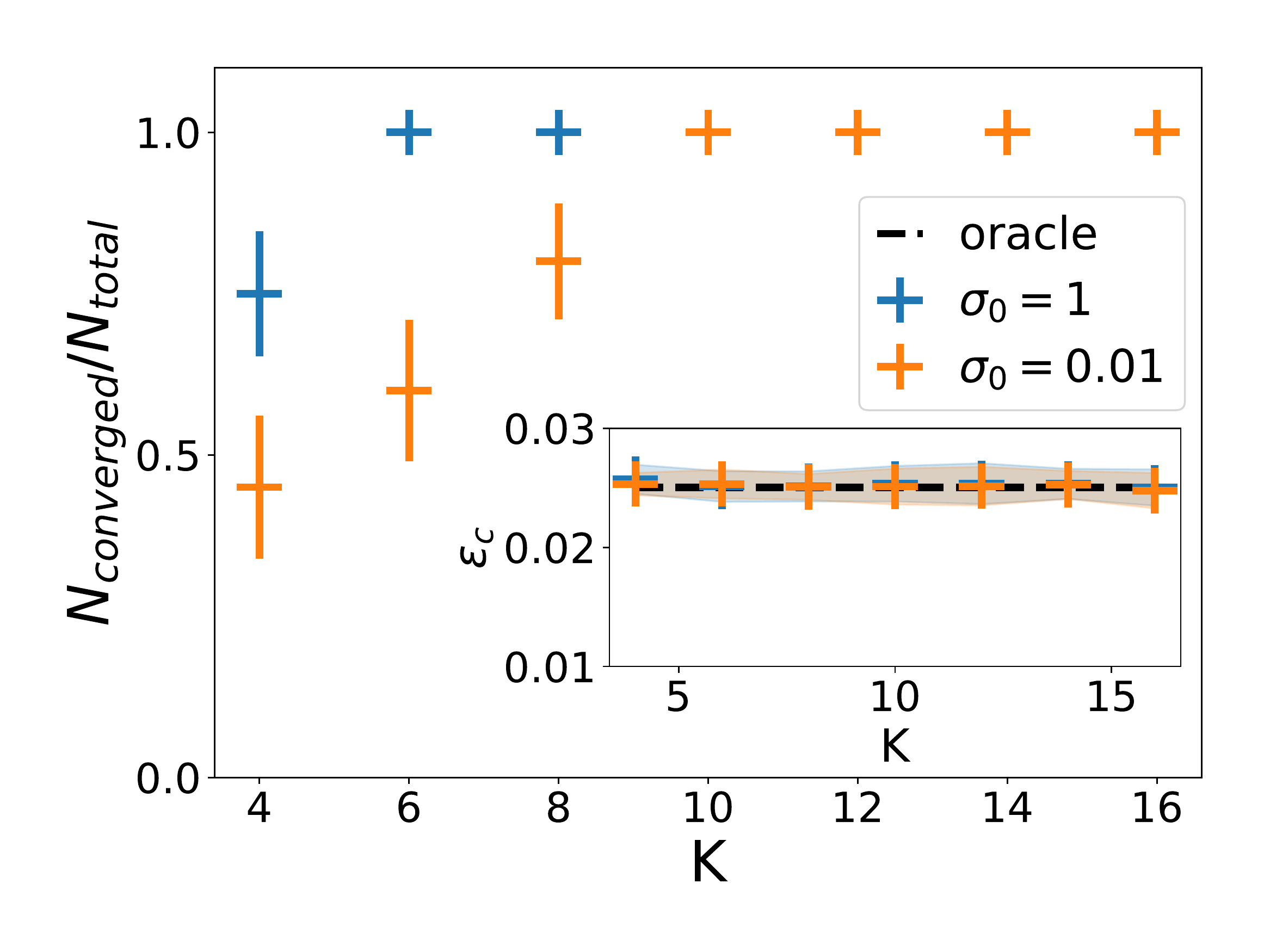}
  % \vspace{-5mm}
  \vspace{-0.5cm}
  \caption{\label{fig:overparametrisation} Fraction of simulations that
    converged to the optimal solution for the 2LNN out of 20 simulations for
    increasing values of $K$. Overparametrisation increases the probability of
    finding the optimal solution but does not affect classification
    performances.  Simulations started with initial weights of
    std.dev. $\sigma_0$. The inset shows the classification error of the
    networks that converged. \emph{Parameters:}
    $D=800, \eta=0.1, \kappa=0, \sigma^2=0.1$, run time $10^5$.  }
    \vspace{-0.4cm}
\end{figure}
\subsection{The impact of over-parametrisation}

We also studied the effect of over-parametrisation, which we define as the number of
additional neurons a student has on top of the $K=4$ neurons that it needs to
reach the oracle's performance on the XOR mixture. We show in the inset
of Fig.~\ref{fig:overparametrisation} that over-parametrisation
does \emph{not} improve final performance, since the remaining error of the student is
dominated by ``spill-over'' of points from one mixture into adjacent
quadrants. However, over-parametrisation leads to an ``implicit acceleration''
effect: over-parametrised networks are much more likely to converge to a
solution that approaches the oracle's performance, as we show in the main of
Fig.~\ref{fig:overparametrisation}.  The term ``implicit acceleration'' was
coined by~\textcite{arora2018optimization} for similar effects in deep neural
networks, and analysed for two-layer networks in the teacher-student setup
by~\textcite{livni2014computational, safran2018spurious}. A complete
understanding of the phenomenon remains an open problem, which we leave for
future work.

\section{Random features on GM classification}
\label{sec:rf_performances}
To understand the performance of random features on Gaussian Mixtures classification, we analyse the performances of the linear model~\eqref{eq:rf_phi} trained with online SGD with the squared loss on the random features $z$~\eqref{eq:rf_z}~\cite{steinwart2009optimal, caponnetto2007optimal}.

First, we assume that we have enough samples, so that $N \gg P$, and discuss the
situation when~$N~\ll~P$ later. For any finite $D,P$, running
the algorithm up to convergence then corresponds to taking the limit~$t\to\infty$. The random features' weights converge to an estimate $\hat{W} $ which can be
computed analytically, see Eq.~\ref{eq:asymptotic_rf_solution}, and allows to
precisely characterise the test error:
\begin{equation}
  \label{eq:rf_asymptotic_error}
  \pmse_{t\to\infty} = \frac{1}{2}\Big( 1 - \sum_\tau\frac{\tilde \Phi_\tau^2}{\rho_\tau} \Big),
 \end{equation}
 where $\rho_\tau$ are the eigenvalues of the feature's covariance matrix
 $\Omega_{ij} = \EE z_i z_j$, with associated eigenvector~$\Gamma_\tau$.
 $\tilde \Phi_\tau \!\equiv\! \sum_{i=1}^P \Gamma_{\tau i} \Phi_i / \sqP$ is the
 input-label covariance after rotation into the eigenbasis of $\Omega$ (see
 Appendix~\ref{app:rf_error}).  Crucially, the test error and $\hat{W} $ \emph{only}
 depend on the first two moments of the features.  The formula for these
 moments, as well as the one for the classification error can be obtained when
 $P,D$ are large using the Gaussian equivalence of~\cite{goldt2020gaussian}. Indeed, the distribution of the
 features $z$ remains a mixture of distributions (see App.~\ref{app:rf}). We
 then define
 \begin{equation}
   M_\alpha\!=\!\sum_{i=1}^P\frac{\hat{w} _i \mathop{\EE}_\alpha[z_i]}{\sqP},
  \text{ }
 Q_\alpha\!=\!\sum_{i=1}^P\frac{\hat{w} _i \hat{w} _j}{P}\mathop{\cov}_\alpha(z,z),
\end{equation}
and we find for $D,P$ large enough, that
\begin{equation}
  \label{eq:rf_asymptotic_classification_error}
    \eclass{t\to\infty} = \frac{1}{2}\left(1-\sum_\alpha \mathcal{P}_\alpha \, y \, \erf\left(\frac{M_\alpha}{\sqrt{2 Q_\alpha}}\right) \right). 
\end{equation}
As discussed in App.~\ref{app:relu_features}, in the case of ReLU activation
function, the feature distribution $p(z_i)$ is a truncated Gaussian. Hence, at
large $D$, $P$, both the mean of $z_i$ and the population covariance
$\cov (z_i,z_j)$ can be obtained analytically in terms of the matrix $F$ and
means $\mu$, see Eq.~\eqref{eq:app:relu_muZ} and~\eqref{eq:app:relu_covZ} for
the full result.

\begin{figure}[t!]
  \centering
  \includegraphics[width=7.5cm]{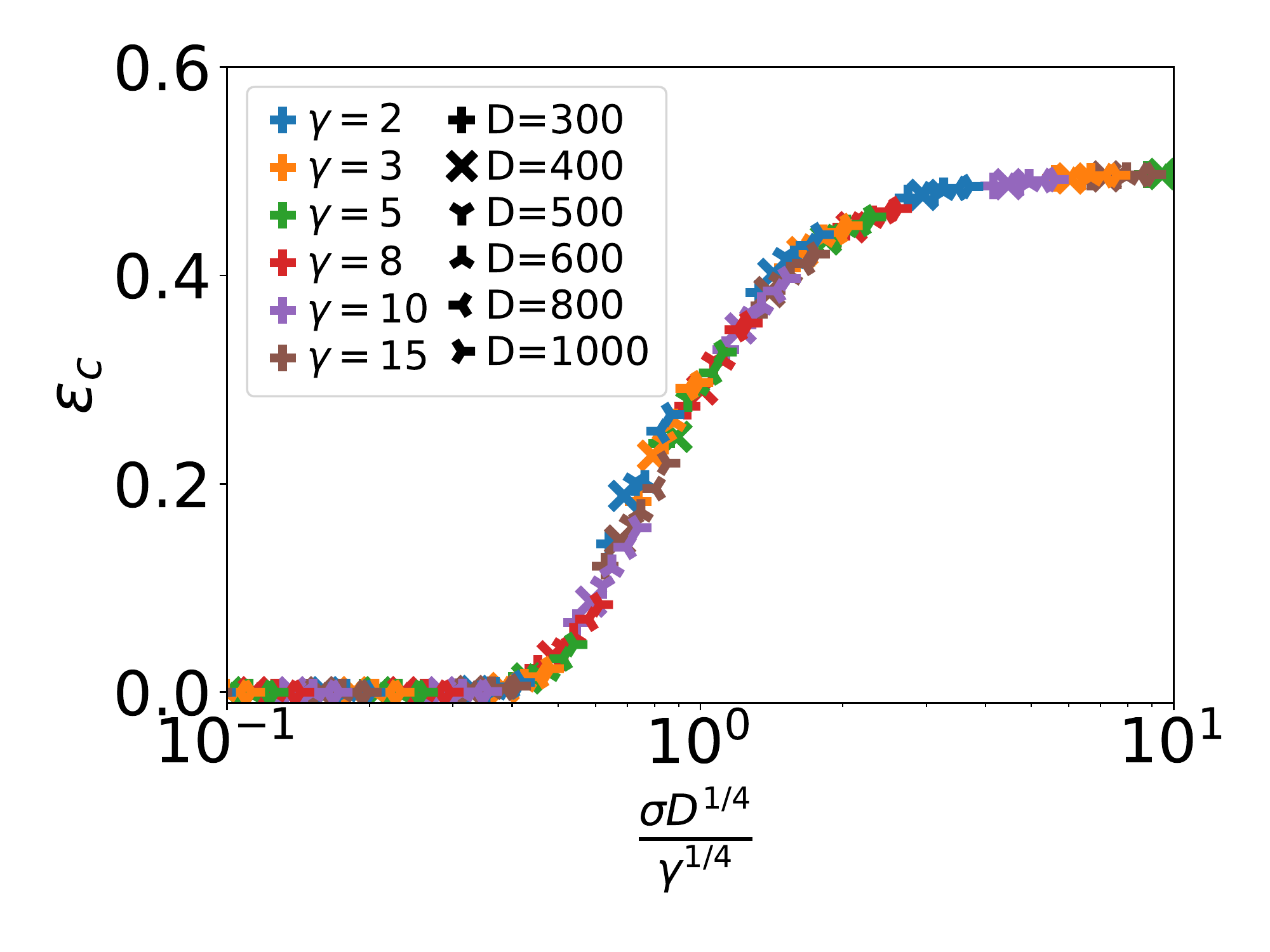}
  \vspace{-0.5cm}
  \caption{\label{fig:snr_manygamma} \textbf{Evolution of the classification
      error of random features (RF) for various values of~$\sigma$,~$\gamma=P/D$
      and~$D$} on the XOR mixture of Fig.~\ref{fig:snr_error}, in the limit of
    large number of samples $N \gg P$. All these different cases can be
    collapsed into a single master curve by plotting the classification
    error versus~$\sigma D^{1/2}/P^{1/4} =\sigma D^{1/4}/\gamma^{1/4}$, showing that
    for large $D,P$, it should be a function
    $\epsilon_c = f(\sigma D^{1/2}/P^{1/4})$.  For a finite input dimension $D$,
    increasing the number of features allows RF to perform increasingly better
    as they approach the Kernel limit. Analytical predictions are obtained by
    the linear regression analysis of
    Eq.~\eqref{eq:rf_asymptotic_classification_error} for
    RF. \emph{Parameters:}~$\eta=0.1$, $\nicefrac{|\mu|}{\sqD}=1$. Note however
    that this analysis requires $N \gg P$. Given random features are sensitive
    to the minimum of $P$ and $N$~\cite{ghorbani2019limitations,
      mei2021generalization} the effective scaling variable is rather
    $\sigma D^{1/2}/\min(P,N)^{1/4}$ (see text). }
    \vspace{-0.4cm}
\end{figure}

We used this formula to obtain precisely the
error~\eqref{eq:rf_asymptotic_error}, and the results are shown in
Fig.~\ref{fig:snr_manygamma}.  We see that the RF error is a function of
$ \sigma D^{1/2}/P^{1/4}=\sigma(\nicefrac{D}{\gamma})^{1/4}$
leading to the conclusion that -- as discussed in Fig.~\ref{fig:snr_error} -- the
``transition'' from the high to low $\snr$ regime happens when
$\sigma^{-1} ~\approx ~\nicefrac{{D}^{1/2}}{P^{1/4}}$.  This scaling further
reveals that $P \approx D^2$ features are required in order to obtain good
performance. The validity of Eq.~\ref{eq:rf_asymptotic_classification_error} is verified in Fig.~\ref{fig:numerical_agreement_snr}. Reaching this performance, however, requires the number of samples $N$ to be larger than $P$, so
$N > O(D^2)$. In the so-called \emph{high-dimensional} regime analysed in this
paper, where $N \approx D$, such performances remain out of reach.  The scaling
analysis can be easily generalised; as discussed
by~\textcite{ghorbani2019limitations, mei2021generalization}, the relevant dimension
for RF performances is, rather than $P$, the minimum between $N$ and $P$.

The classification error is thus a function of
$\sigma\nicefrac{D^{1/2}}{\min(N,P)}^{1/4}$. If $N$ is $O(D)$,
then even in the kernel limit when $P \to \infty$, the performance degrades to
no more than a random guess as soon as
\begin{equation}
\sigma \gg  \nicefrac{N^{1/4}}{D^{1/2}},
%% so that the RHS goes to 0 when D-> infty so even for sigma very small you do no better than chance
\end{equation}
and therefore for any value of $\sigma$ when $D,N \to \infty$ with fixed $N/D$. In a nutshell, for any fixed $\sigma$, lazy training methods such as random features or kernels will fail to beat a random guess in the high-dimensional limit.
This, and the requirement of at least $N=O(D^2)$ samples to learn, are to be contrasted with the the oracle-like performance achieved by a simple neural net with only $N=O(D)$ samples.

\balance
\paragraph{Why do random features fail? The linear regime of features maps}

This raises the question of \emph{why} the random feature and kernel methods
fail in the high-dimensional setting. As we shall
see -- and this has been already discussed in different contexts by
\textcite{elkaroui2010spectrum, mei2019generalization} -- this can be understood
analytically from the fact that the feature map is effectively linear when when
$P=O(D)$. This section is now dedicated to computing the moments
$\mathop{\EE}_\alpha[z_i]$ and $\mathop{\cov}_\alpha(z,z)$ analytically in this
region, revealing that this effective linearity is indeed the underlying reason for the failure of RF in this
regime.

Since the mixture remains a mixture after the application of random features, our main task is to compute the new means and variances of the distribution in the transformed space.  We thus focus on transformation of a random variable drawn from a single Gaussian $x_r =  \mu_r / \sqrt{D} + \sigma w_r$, 
%\begin{equation}
%  \label{eq:1}
%  x_r = \frac{  \mu_r}{\sqrt{D}} + \sigma w_r , 
%\end{equation}
where $w_r$ is a standard Gaussian, in the kernel, or random feature, space. 

For generic activation function the first two moments of the features can be obtained in the well studied low signal-to-noise regime $\snr\sim O(1)$.
Key to do so, is the observation that $\nicefrac{ F_{ir} \mu_r}{D}\sim O(\nicefrac{1}{\sqD})$. The activation function can thus be expanded in orders of $\nicefrac{1}{D}$ and its action is essentially linear. 
We define the constants 
\begin{equation}
\label{eq:abc}
    a\equiv \EE\psi\left( \sigma \zeta \right),\quad b\equiv\EE\zeta\psi\left( \sigma \zeta
  \right), \quad
  c^2\equiv\EE \psi\left( \sigma \zeta \right)^2
\end{equation}
%\begin{gather}
%  \label{eq:abc}
%  a\equiv \EE\psi\left( \sigma \zeta \right), %\text{ } b\equiv\EE\zeta\psi\left( \sigma \zeta
%  \right), \text{ } c^2\equiv\EE \psi\left( %\sigma \zeta \right)^2,
%\end{gather}
with the expectation taken over the standard Gaussian random variable $\zeta$.
To leading order, the mean and covariance of the features are given by (cf. Sec.~\ref{app:rf}):
\begin{align}
  \EE z_i &= a +  b \sum_{r=1}^D\frac{F_{ir} \mu_r }{\sigma D}\label{eq:rf_low-snr_mean}\, \\ 
  \operatorname{cov}(z_i,z_j)&=
                                       \begin{cases}
                                         c^2 - a^2,
                                         \quad & i =j \, ,\\
                                         b^2 \sum_{r}\frac{F_{ir}F_{jr}}{D}\quad &
                                         i\neq j.
    \end{cases}\label{eq:rf_low-snr_cov}
\end{align}

This computation immediately reveals the reason random features cannot hope to learn in the low $\snr$ regime: the transformation of the
means is only linear; hence a Gaussian mixture that is not linearly separable in input space will remain so even after random features.
In other words, if the centres of the Gaussian are too close, the kernel fails to map the data non-linearly to a large dimensional space. In contrast, in the high-$\snr$ regime, where the centres are separated enough, the non-linearity kicks-in and the data becomes separable in feature space.
\balance

\begin{figure*}[t!]
  \centering
  \includegraphics[width=\linewidth]{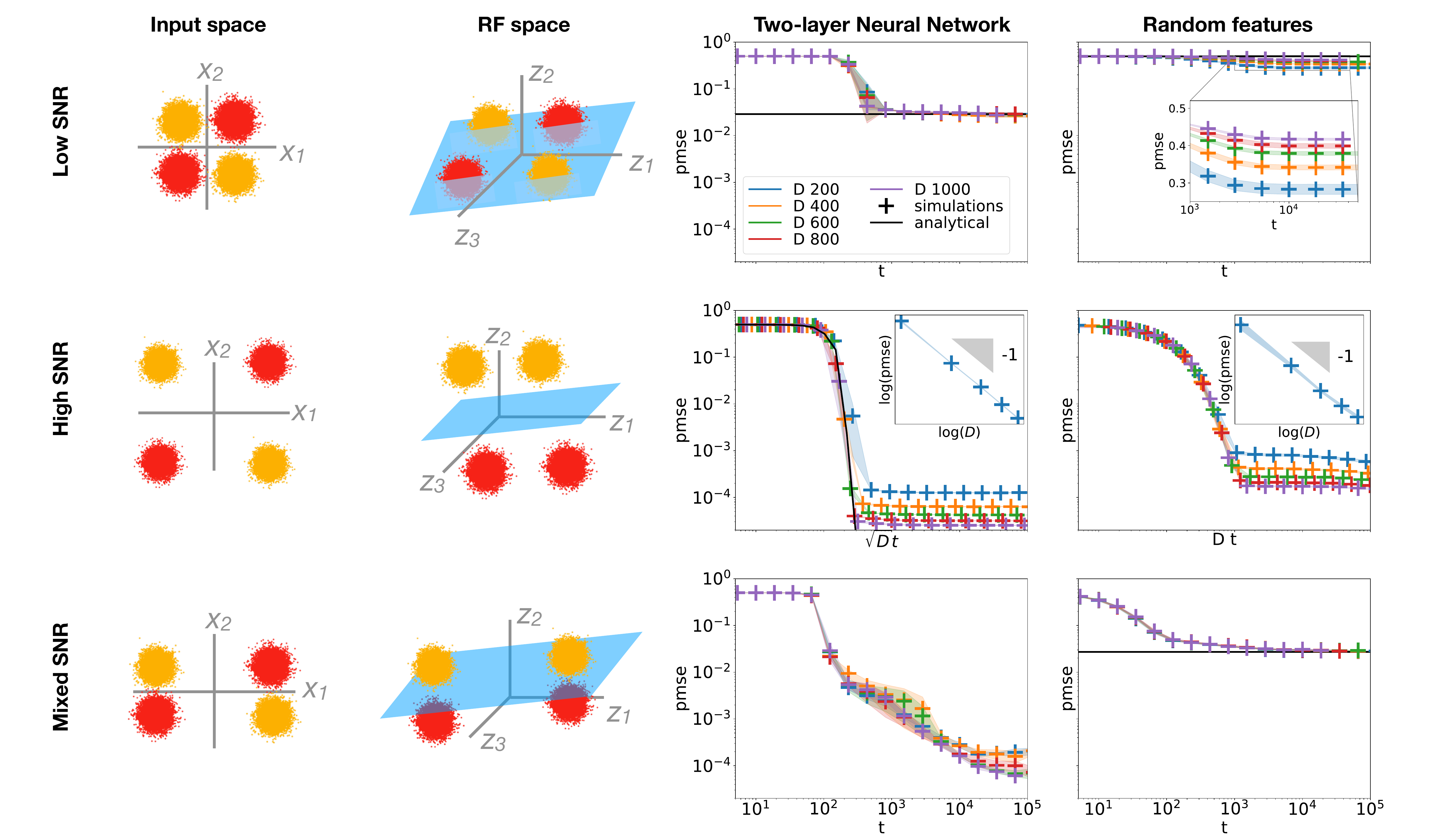}
\vspace{-0.5cm}
  \caption{\label{fig:xor} \textbf{The performance of 2LNN and RF on the
      XOR-like mixture with different signal-to-noise ratios.} The \emph{first
      and second columns} show the XOR-like mixture in input space and random
    feature space, resp. The \emph{third and fourth columns} show the
    $\pmse$~\eqref{eq:pmse} of 2LNN and RF during training, resp. In the low $\snr$
    regime (top row), while the 2LNN learns a non trivial function of the
    inputs, RF cannot perform better than random chance since the XOR-like
    mixture remains one in feature space. Both networks learn to classify the
    XOR-mixture in the high $\snr$ regime (middle row) as the clusters become well
    separated in feature space. For mixed $\snr$ (bottom row), even though both
    networks do better than random guessing, 2LNN outperform RF as the distance
    between opposite sign clusters remains of order one in feature space.  In
    all plots, crosses are obtained from simulations with input dimension
    $D=1000$. Solid black lines in the 2LNN plots are obtained using tools from
    Sec.~\ref{sec:odes} (long-time performance of Sec.~\ref{sec:state-evolution}
    for the first row and integration of the odes for the second). Solid black
    lines in the RF plots indicate the test error obtained from the analysis of
    Sec.~\ref{sec:rf_performances} with $D=10000$.
    $\sigma^2=0.05, P=2D, \eta=0.1, K=10, \sigma_0=10^{-2}$.}
   \vspace*{-1em}
\end{figure*}

\paragraph{Relation to kernel methods}
The same argument explains the failure of kernel methods: if two centres $\bf x$ and $\bf y$ are close, the kernel function $K({\bf x},{\bf y})$ can be expanded to low order and the kernel is essentially linear,  leading to bad performances. The connection can be made explicit using the convergence of random features to a kernel~\cite{rahimi2008random, rahimi2009weighted}: 
\begin{equation}
  K(\boldsymbol x, \boldsymbol y)\!=\!\frac{1}{P}\!\sum_{i=1}^P
  \mathop{\EE}_F\!\left[\psi\left(\sum_{r=1}^D\frac{x_r F_{ir}}{\sqrt
      D}\right)\psi\left(\sum_{s=1}^D\frac{y_s F_{is}}{\sqrt D}\right)\right],
\end{equation}
At low SNR, the constants $a, b, c$ can be obtained from the kernel via
\begin{gather}
  \begin{aligned}
    c^2=\EE K(\sigma\boldsymbol \omega_1,\sigma\boldsymbol \omega_1),
    \quad a^2 = \EE K(\sigma\boldsymbol \omega_1,\sigma\boldsymbol \omega_2),\\
    b^2=D \sigma^2\left[ -a^2 + \EE
    K\left(\frac{\boldsymbol \mu}{\sqrt{D}}+\sigma\boldsymbol \omega_1,\frac{\boldsymbol \mu}{\sqrt{D}}+\sigma\boldsymbol \omega_2\right)\right] \nonumber
  \end{aligned}
\end{gather}
where the average is taken over two standard Gaussian random vectors $\omega_1, \omega_2\in\mathbb R^D$. 
This relation, similar in nature to one of \textcite{elkaroui2010spectrum}, allows to express the statistical properties of the features directly from the kernel function.

\section{Neural networks vs random~features}
\label{sec:xor}

We now collect our results for a comparison of the performance of
2LNN and RF on the XOR-like mixture from Fig.~\ref{fig:snr_error}. We look at three
different regimes for the $\snr$, illustrated in the first column of
Fig.~\ref{fig:xor}. The second column visualises the mixture after the
Gaussian random features transformation with $\psi(x)=\max(0, x)$~\eqref{eq:rf_z}. The third and
fourth columns show the evolution of the $\pmse$ of 2LNN
and RF, respectively, during training with online SGD. Since overparametrisation does not impact the 2LNN's performance in these tasks, Sec.~\ref{sec:state-evolution}, we train a $K=10$ network to increase the number of runs that converge.

At \textbf{low $\snr$} (a) the distance of each Gaussian to the origin is $O(1)$
and the standard deviation $\sigma\sim O(1)$ as well. The two-layer neural
network learns to predict the correct labels almost as well as the
oracle~\eqref{eq:ec_oracle}. Its performance does not depend on $D$, and using the long-time solution of
Sec.~\ref{sec:state-evolution}, we can predict its asymptotic error (black line) which agrees well with simulations (crosses).
In contrast, random features display an asymptotic
error that approaches random guessing as the input dimensions increases
(inset). This is clear from Eq.~\ref{eq:rf_low-snr_cov}: in the large $ D$
limit, random features only produce a \emph{linear} transformation of their
input. The XOR therefore remains a XOR in RF space, leading linear regression's failure to do better than chance.

At \textbf{high $\snr$} (b), the distance between the clusters scales as $\sqrt{D}$
while the $\sigma$ remains fixed. The asymptotic error of
the 2LNN thus decreases with $D$ and the network is able to learn perfectly in the
$D\to\infty$ limit (black line).  The error of random features also approaches~0
as $D\to\infty$, since the mixture is now well separated in random feature
space, too.

We finally consider a regime of \textbf{mixed $\snr$} (c) where the mixture is
well-separated in one dimension, but very close in the other dimension. We
achieve this by setting $\mu^0_1\!~\sim~\!\sqD, \mu^0_2~\!\sim\!~D$ for the mean
of the first mixture, etc. Random features then achieve a non-trivial
generalisation error, which can be understood by considering the means of the
features $z_i$. 
The large component $\mu_2$, induces the activation function to perform a non-linear transformation of the centres and allows for opposite sign centroids to be separated by a
hyper-plane in feature space. The small component $\mu_1$, causes the distance between opposite sign centroids, which is of order $O(1)$ in input space, to remain of order one in feature space, for all $D$. This leads to a finite generalisation error of RF which
remains invariant with increasing input dimension. In this regime, the 2LNN still achieve better performance than the random features, thereby completing the
picture we developed in Fig.~\ref{fig:snr_error}.

\balance

\section*{Acknowledgements}
We acknowledge funding from the
ERC under the European Union’s Horizon 2020 Research and Innovation Programme
Grant Agreement 714608-SMiLe, from ``Chaire de recherche sur les modèles et
sciences des données'', and from the French
National Research Agency grants ANR-17-CE23-0023-01 PAIL and ANR-19-P3IA-0001 PRAIRIE.

\bibliography{nn}
\bibliographystyle{icml2021}

\balance

\newpage
\appendix
%%%%%%%%%%%%%%%%%%%%%%%%%%%%%%%%%%%%%%%%%%%%%%%%%%%%%%%%%%%%%%%%%
\onecolumn

%%%%%%%%%%%%%%
\numberwithin{equation}{section}% \renewcommand{\theequation}{S.\arabic{equation}}

\section{Summary of Notations}
\begin{center}
  \def\arraystretch{2}
%\begin{tabular}{p{ 2cm}  p{5.5cm} p{2cm} p{5.5cm}}
\begin{tabular}{p{.2\linewidth}  p{.25\linewidth} p{.2\linewidth}  p{.25\linewidth}}
$D$ & input dimensions &
$P$ & number of random features \\
$N$ & number of samples &
$\gamma=\nicefrac{P}{D}$ &   \\
$t=\nicefrac{N}{D}$  & training time, or equivalently rescaled number of training samples  &
$y$ & true label \\
$\Omega^\alpha\in\mathbb{R}^{D\cross D}$ & covariance of the normal distribution of cluster $\alpha$&
$\frac{\mu^\alpha}{\sqD}\in\mathbb{R}^D$ & mean of the Gaussian cluster $\alpha$ in the mixture\\
$x\in\mathbb{R}^D$ & input &
$\sigma$ & standard deviation of the Gaussian clusters in a mixture with $\Omega_{\alpha}=\sigma^2\mathbb{I}$\\
$\snr\!=\!\frac{|\mu|}{\sigma \sqD}$ & signal to noise ration &
$\eta$ & learning rate\\
$\kappa$ & $L2$-regularisation constant&
$\pmse$ & population mean squared error\\
$\eclass{ } $& classification error 
\end{tabular}
\end{center}
$\text{ }q(x|y)\!=\!\!\sum\limits_{\alpha\in\mathcal S(y)}\!P_{\alpha^\pm}\mathcal{N}_\alpha(x)$\quad conditional probability of $x$ given the true label $y$

\paragraph{Two layer neural networks (2LNN)}
\begin{center}
  \def\arraystretch{2}
\begin{tabular}{p{.2\linewidth}  p{.25\linewidth} p{.2\linewidth}  p{.25\linewidth}}
$K$ & number of hidden nodes of the 2LNN&
$w_i^k$& first layer weights\\
$v^k$& second layer weights&
$g:\mathbb{R}\to\mathbb{R}$ & activation function \\
$\lambda\equiv\sum\limits_r\nicefrac{w_r^k x_r}{\sqD}$ & local field/pre-activation of the 2LNN &
$\phi(\theta)= \sum\limits_{k=1}^K v^k g(\lambda^k)$ & output of the network \\
$Q_\alpha^{kl}\!=\!\sum\limits_{rs}\frac{w^k_r\Omega^\alpha_{rs}w^l_s}{D}$ & order parameter/ covariance of the local fields &
$ M_\alpha^{k}\!=\!\sum\limits_{r}\frac{w^k_r\mu_r^\alpha}{D}$ & order parameter/mean of the local fields\\
$\sigma_0$ & weights are initialised i.i.d.~from $\mathcal{N}(0,\sigma_0^2)$ &
\end{tabular}
\end{center}

\newpage
\paragraph{Random Features (RF)}

\begin{center}
  \def\arraystretch{2}
\begin{tabular}{p{.2\linewidth}  p{.2\linewidth} p{.25\linewidth}  p{.25\linewidth}}
 $ F\in\mathbb{R}^{P \times D} $   & projection matrix \qquad\qquad $F_{ir}\!\overset{\text{i.i.d.}}{\sim}\!\mathcal{N}(0,1)$ &
 $z\equiv \psi\left( \sum\limits_{r=1}^D\nicefrac{F_{ir}x_r}{\sqD}\right)$ &  features\\ 
 $\psi:\mathbb{R}\to\mathbb{R} $ & activation function applied element wise & 
 $\phi(\theta)= \sum\limits_{i=1}^P \nicefrac{w_i z_i}{\sqD} $              &  output of the network \\
 $\hat{W}$ & fix point solution of the SGD update equation of RF
\end{tabular}
\end{center}
\section{Derivation of the dynamical equations}
\label{app:ODEs}

In this appendix, we derive the dynamical equations that describe the dynamics of two-layer neural networks trained on the Gaussian mixture from Sec.~\ref{sec:odes}. We first derive a useful Lemma for the averages of weakly correlated random variables~\ref{app:generalised_gep}, which we we then use in the derivation of the dynamical equations~\ref{app:derivODEs}

\subsection{Moments of functions of weakly correlated variables}
\label{app:generalised_gep}
Here, we show how to compute expectation of functions of weakly correlated
variables with non zero mean. The derivation follows the ones of~\cite{goldt2020modelling} (see App.~A). We extend their computations to include
variables with non-zero means.

Consider the random variables $x$,$y\in\mathbb{R}$, jointly Gaussian with joint probability distribution:
\begin{equation}
    P(x, y)=\frac{1}{2\pi \sqrt{\det M_{2}}} \exp \left[-\frac{1}{2}\left(\begin{array}{ll}
x-\bar x & y-\bar y
\end{array}\right) M_2^{-1}\left(\begin{array}{l}
x-\bar x \\
y-\bar y
\end{array}\right)\right],
\label{eq:joint_gaussian}
\end{equation}
where we defined the mean of $x$, respectively $y$, as $\bar x$, respectively $\bar y$ and the covariance matrix:
\begin{align}
    &M_2=\left(\begin{array}{ll}C_x & \epsilon M_{12}\\\epsilon M_{12}&C_y\end{array}\right).
\end{align}
The weak correlation between $x$ and $y$ is encapsulated in the parameter $\epsilon \ll 1$ while $M_{12}\!\sim\!O(1)$.

We are interested in computing expectations of the form $\displaystyle\mathop{\EE}_{(x,y)} [f(x)g(y)]$ with two real valued functions $f,g: \mathbb{R}\to\mathbb{R}$. Leveraging the weak correlation between $x$ and $y$, we can expand the distribution Eq.~\eqref{eq:joint_gaussian} to linear order in $\epsilon$, i.e.:
\begin{align}
   (\eqref{eq:joint_gaussian})= \frac{1}{2\pi\sqrt{C_x C_y}}\operatorname{e}^{-\frac{1}{2 C_x}(x-\bar x)^2-\frac{1}{2 C_y}(y-\bar y)^2 }{\left[1-\epsilon (x-\bar x) \left(C_x^{-1} M_{12} C_y^{-1}\right) (y-\bar y) +O\left(\epsilon^2\right)\right]}
\end{align}
Using the above, one can compute the expectations:
\begin{align}
\label{eq:2pt}
    \mathop{\EE}_{(x,y)}\left[f(x)g(y) \right]=&\mathop{\EE}_{x}\left[f(x)\right]\mathop{\EE}_{y}\left[g(y)\right]\notag\\
    &+\epsilon\mathop{\EE}_{x}\left[f(x)(x-\bar x)\right]\left(C_x^{-1} M_{12} C_y^{-1}\right)\mathop{\EE}_{y}\left[g(y)(y-\bar y)\right]+O(\epsilon^2).
\end{align}
The expectations are now taken over the 1-dimensional distributions of $x\!\sim\!\mathcal{N}(0,C_x)$ and $y\!\sim\!\mathcal{N}(0,C_x)$.

Similarly, consider the case of three weakly correlated real random variables
$x_i$, $i=1,2,3$ with mean $\bar x_i$ and covariance matrix $M_3$ such that
\begin{equation}
    (M_{3})_{ij}=\begin{cases}
    C_i, \text{ if } i = j \\
    \epsilon M_{ij}, \text{ if }i \neq j
         \end{cases}.
\end{equation}
One can use an expansion of the joint probability distribution of $\{x_i\}$ to linear order in $\epsilon$ to compute three point moments of real valued functions $f,g,h$ as:
\begin{align}
\mathop{\EE}_{\{x\}_i}\left[f(x_1)g(x_2)h(x_3)\right]=&\mathop{\EE}_{\{x\}_i}\left[f(x_1)\right]\left[g(x_2)\right]\left[h(x_3)\right]\notag\\&-\epsilon\mathop{\EE}_{x_3}[h(x_3)]\mathop{\EE}_{x_1}[(x_1-\bar x_1) f(x_1)]\left(C_x^{-1} M_{12} C_y^{-1}\right)\mathop{\EE}_{x_2}[(x_2-\bar x_2) g(x_2)]\notag\\
    &-\epsilon\mathop{\EE}_{x_2}[g(x_2)]\mathop{\EE}_{x_1}[(x_1-\bar x_1) f(x_1)]\left(C_x^{-1} M_{13} C_y^{-1}\right)\mathop{\EE}_{x_3}[(x_3-\bar x_3) h(x_3)]\notag\\
    &-\epsilon\mathop{\EE}_{x_1}[f(x_1)]\mathop{\EE}_{x_2}[(x_2-\bar x_2) g(x_2)] \left(C_x^{-1} M_{23} C_y^{-1}\right)\mathop{\EE}_{x_3}[(x_3-\bar x_3) h(x_3)]+O(\epsilon^2).
\end{align}
In the case in which $x_1$ and $x_2$ are weakly correlated with $x_3$ but not between each other, i.e. $\operatorname{Cov}(x_1,x_2)=M_{12}\!\sim\!O(1)$, one has:
\begin{align}
\begin{split}
    \EE\left[f(x_1)g(x_2)h(x_3) \right] =&\EE\left[f(x_1)g(x_2) \right]\EE\left[h(x_3) \right]\\
     &+\epsilon\frac{\EE\left[h(x_3)(x_3-\bar x_3) \right]}{ \left(C_{x_1}C_{x_2}-M_{12}^2\right) C_{x_3}}\left\{ \EE\left[f(x_1)g(x_2)(x_1-\bar x_1)\right]M_{13} C_{x_2}\right.\\
     &\hspace{11em}\left.+\EE\left[f(x_1)g(x_2)(x_2-\bar x_2)\right]M_{23}C_{x_1} \right. \\
     &\hspace{11em}\left.-\EE\left[f(x_1)g(x_2)(x_1-\bar x_1)\right]M_{12}M_{23}\right.\\
      &\hspace{11em}\left.-\EE\left[f(x_1)g(x_2)(x_2-\bar x_2)\right]M_{13}M_{12}\right\}+O(\epsilon^2).
\end{split}
\end{align}

\subsection{Derivation of the ODEs}
\label{app:derivODEs}
In this section, we derive the ODEs describing the dynamics of training of a 2LNN trained on inputs sampled from the distribution~\eqref{eq:data} with $L2$-regularisation constant $\kappa$. We restrict to the case where all the Gaussian clusters have the same covariance matrix, i.e.~$\Omega^\alpha=\Omega\in\mathbb{R}^{D\cross D}$.

In order to track the training dynamics, we analyse the evolution of the macroscopic operators defined in Eq.~\eqref{eq:OP} allowing to compute the performances of the network at all training times.

At the $s$th step of training, the SGD update for the networks parameter is
given by Eq.~\eqref{eq:sgd}:
\begin{align}
  \dd w_{i}^k \equiv \left(w_{i}^k\right)_{s+1}- \left(w_{i}^k\right)_{s}
  =-\frac{\eta_{W}}{\sqrt{D}}\left( v^k \Delta g'(\lambda^k) x_i +  \kappa w^k_i\right) , \qquad
  \dd v^k = - \frac{\eta_{v}}{\sqrt{D}} \left( g(\lambda^k)+  \kappa v^k \Delta\right).
\end{align}
In order to guarantee that the dynamics can be described by a set of ordinary equations in the $D\to\infty$ limit, we choose different scalings for the first and second layer learning rates:
$$\eta_{W}=\eta,\qquad \eta_{v}=\eta/D$$ for some constant $\eta$.

\paragraph{Update of the first layer weights}
To make progress, consider the eigen-decomposition of the covariance matrix:
\begin{align}
    \Omega_{rs}&=\frac{1}{D}\sum_{\tau=1}^D \Gamma_{s\tau}\Gamma_{r\tau}\rho_\tau;
\end{align}
where we denote the eigenvalues as $\rho_\tau$, their corresponding eigenvector as $\Gamma_\tau$ and the eigenvalue distribution as $p_\Omega$. We further define the projection of the weights into the projected basis as
\begin{equation}
    \tilde w_\tau^k \equiv \frac{1}{\sqD}\sum_{\tau=1}^D \Gamma_{s\tau} w^k_s
\end{equation}
and similarly $\tilde x_\tau$ and $\tilde \mu^\alpha_\tau$ as the projected inputs and means. In this basis, the SGD update for the first layer weights is:
\begin{align}
  \dd \tilde w_{\tau}^k \equiv \left(\tilde  w_{\tau}^k\right)_{s+1}- \left(\tilde w_{\tau}^k\right)_{s}
  =-\frac{\eta}{\sqrt{D}} \left(v^k \Delta g'(\lambda^k) \tilde x_\tau+\kappa w_{\tau}^k \right).
\end{align}

The expectation of this update over the distribution Eq.~\eqref{eq:data} is given by:
\begin{align}
    \EE \dd \tilde{w}_\tau^k =\sum_{\alpha\in\mathcal{S}\left(+\right)} \mathcal{P}_\alpha  \dd \tilde w^{k}(\rho)_{\alpha^+} +\sum_{\alpha\in\mathcal{S}\left(-\right)} \mathcal{P}_\alpha\dd \tilde w^{k}(\rho)_{\alpha^-},
\end{align}
where we decomposed the expectation into the different clusters and introduced:
\begin{align}
  \dd \tilde w^{k}(\rho)_{\alpha^\pm}
  =& \pm \frac{\eta}{\sqrt{D}} v^k \mathcal{C}^k_\tau -\frac{\eta}{\sqrt{D}} \sum_{j\neq k}v^k v^j \mathcal{A}^{kj}_\tau -\frac{\eta}{\sqrt{D}} v^k v^k \mathcal{B}^k_\tau- \frac{\eta \kappa}{\sqrt{D}}\tilde w^{k}(\rho),
  \label{eq:delta_wtau_alpha}
\end{align}
with the expectations $\mathcal{A}^{kj}_\tau, \mathcal{B}^k_\tau$ and $\mathcal{C}^k_\tau$ defined as:
\begin{equation}
\label{eq:ABC}
   \mathcal{A}^{kj}_\tau = \mathop{\EE}_{\alpha} g^\prime(\lambda^k)g(\lambda^j)\tilde x_\tau,\quad \mathcal{B}^k_\tau= \mathop{\EE}_{\alpha} g^\prime(\lambda^k)g(\lambda^k)\tilde x_\tau,\quad \mathcal{C}^k_\tau=\mathop{\EE}_{\alpha} g^\prime(\lambda^k)\tilde x_\tau.
\end{equation}

A crucial observation, is that $\lambda^k$ and the projected input
$\tilde x_\tau$ are jointly Gaussian and weakly correlated, with a correlation of order
$1/\sqrt{D}$:
\begin{equation}
\mathop{\EE}_{\alpha} \lambda^k \tilde x_\tau = \frac{1}{\sqrt D}
\rho_\tau\tilde w^k.
\end{equation}
Thus, we can compute the expressions Eq.~\eqref{eq:ABC} using the proposition for
weakly correlated variables derived in App.~\ref{app:generalised_gep}. This
gives:
\begin{align}
    \label{eq:averages_first_layer}
. \begin{split}
    \mathcal{A}^{kj}_\tau&= \mathop{\EE}_{\alpha} g^\prime(\lambda^k)g(\lambda^j) \frac{\tilde \mu^\alpha_\tau}{ \sqrt{D}}\\
    & \quad +\frac{1}{Q^{kk}Q^{jj}-Q^{kj2}}\left\{\left( \mathop{\EE}_{\alpha} g^\prime(\lambda^k)\lambda^k g(\lambda^j) - \mathop{\EE}_{\alpha} g^\prime(\lambda^k)g(\lambda^j)  M^{\alpha k} \right)\left(Q^{jj}\frac{\tilde{w}^k_{\tau} \rho_\tau}{ \sqrt{D}}-Q^{kj}\frac{\tilde{w}^j_\tau \rho_\tau}{ \sqrt{D}}\right)\right.\\
    & \hspace*{10em} \left.\left( \mathop{\EE}_{\alpha} g^\prime(\lambda^k)\lambda^j g(\lambda^j) - \mathop{\EE}_{\alpha} g^\prime(\lambda^k)g(\lambda^j)  M^{\alpha j} \right)\left(Q^{kk}\frac{\tilde{w}^j_\tau \rho_\tau}{ \sqrt{D}}-Q^{kj}\frac{\tilde{w}^k_{\tau} \rho_\tau}{ \sqrt{D}}\right)\right\}\\
     \mathcal{B}^{k}_\tau&=\mathop{\EE}_{\alpha} g^\prime(\lambda^k)g(\lambda^k) \frac{\tilde \mu^\alpha_\tau}{ \sqrt{D}} + \frac{1}{Q^{kk}}\left( \mathop{\EE}_{\alpha} g^\prime(\lambda^k)\lambda^k g(\lambda^k) - \mathop{\EE}_{\alpha} g^\prime(\lambda^k)g(\lambda^k)  M^{\alpha k} \right)\frac{\tilde{w}^k_{\tau} \rho_\tau}{ \sqrt{D}},\\
    \mathcal{C}^{k}_\tau&=\mathop{\EE}_{\alpha} g^\prime(\lambda^k) \frac{\tilde \mu^\alpha_\tau}{ \sqrt{D}}+\frac{1}{Q^{kk}}\left( \mathop{\EE}_{\alpha} g^\prime(\lambda^k)\lambda^k  - \mathop{\EE}_{\alpha} g^\prime(\lambda^k)  M^{\alpha k} \right)\frac{\tilde{w}^k_{\tau} \rho_\tau}{ \sqrt{D}},
    \end{split}
\end{align}
where we used that the first moments of the local fields are given by the order parameters, $\mathop{\EE}_{\alpha}[\lambda^k]=M^{\alpha k}$ and $\cov(\lambda^k,\lambda^l)=Q^{kl}$.
The multi-dimensional integrals of the activation function only depend on the order parameters at the previous step. We discuss how to obtain them, using monte-carlo methods in Sec~\ref{sec:odes}. The averaged update of the first layer weights follows directly from Eq.~\eqref{eq:averages_first_layer}.

\paragraph{Update of the Order parameters}
In order to derive the update equations for the order parameters, we introduce
the densities $m(\rho,t)$ and $q(\rho,t)$. These depend on $\rho$ and on the
normalised number of steps $t=\nicefrac{\mu}{D}$, which we interpret as a
continuous time variable.
    \begin{align}
    m^{\alpha k}(\rho,t)=\frac{1}{D\epsilon_\rho}  \sum_\tau
  \tilde{w}_\tau^k\tilde{\mu}_\tau^\alpha  \mathbbm{1}_{\left(\rho_\tau \in
    \mathopen[\rho,\rho+\varepsilon_\rho\mathclose[\right)},\notag\\
    q^{kl}(\rho,t)=\frac{1}{D\varepsilon_\rho} \sum_\tau \tilde{w}_\tau^k
  \tilde{w}_\tau^l  \mathbbm{1}_{\left(\rho_\tau \in
    \mathopen[\rho,\rho+\varepsilon_\rho\mathclose[\right)},
    \label{eq:r_tauq_tau}
\end{align}
where $\mathbbm{1}(.)$ is the indicator function and the limit $\epsilon_\rho\to 0$
is taken after the thermodynamic limit. Using these definitions, the order
parameters can be written as:
\begin{align}
    \label{eq:integral_r_q}
    Q^{kl}(t)=\int \dd \rho \;  p_{\Omega}(\rho)  \;\rho q^{kl}(\rho,t),\quad
    M^{\alpha k}(t) =\int \dd \rho \; p_{\Omega}(\rho)  m^{\alpha k}(\rho,t).
\end{align}

The equation of motion of $m$ can  can be easily computed using the update~\eqref{eq:delta_wtau_alpha} and is given by:
\begin{align}
    \frac{\partial m^{\beta k}(\rho,t)}{\partial t} =\sum_{\alpha\in\mathcal{S}\left(+\right)} \mathcal{P}_\alpha \frac{\partial
    m_{\alpha^+}^{\beta k}(\rho,t)}{\partial t} +\sum_{\alpha\in\mathcal{S}\left(-\right)} \mathcal{P}_\alpha\frac{\partial m_{\alpha^-}^{\beta k}(\rho,t)}{\partial t}, 
\end{align}
with
\begin{align}
  \label{eq:eom-m}
  \begin{split}
\frac{\partial m^{\beta k}_{\alpha^\pm}(\rho,t)}{\partial t}=&\pm \eta v^k I_{31}(k)T^{\alpha\beta}\pm\frac{ \eta \rho v^k}{Q^{kk}}\left( I_{32}(k,k)- I_{31}(k) M^{\alpha k} \right)m^{\beta k}(\rho,t)\\
   &  + \sum_{j\neq k} \left[-\eta v^k v^j I_{22}(k,j)T^{\alpha\beta}\right.\\
   &  \qquad\quad +\frac{\eta \rho v^k v^j}{Q^{kk}Q^{jj}-Q^{kj2}}\left\{\left(-I_{3}(k,k,j)+ I_{32}(k,j) M^{\alpha k} \right)\left(Q^{jj}m^{\beta k}(\rho,t)-Q^{kj}m^{\beta j}(\rho,t)\right)\right.\\
       & \hspace{12em}+\left.\left.\!\!\left(- I_3(k,j,j)+ I_{32}(k,j) M^{\alpha j} \right)\left(Q^{kk}m^{\beta j}(\rho,t)-Q^{kj}m^{\beta k}(\rho,t)\right)\right\}\right]\\
       &  - \eta v^k v^k I_{22}(k,k)T^{\alpha\beta}
         -\frac{\eta \rho v^k v^k}{Q^{kk}}\left( I_3(k,k,k)- I_{22}(k,k) M^{\alpha k} \right)m^{\beta k}(\rho,t)-\kappa\eta m^{\beta k }(\rho,t).
    \end{split}
\end{align}
Note how, in order to close the equation, we introduced an additional order parameter $T^{\alpha\beta}= \sum_{r=1}^D\frac{\mu^\alpha_{r}\mu^{\beta}_{r}}{D}$, which is entirely defined by the overlap of the means of the mixture under consideration and is therefore a constant of motion.
For compactness, we defined the multidimensional integrals $I$ of the activation function over the local fields as:
\begin{subequations}
\begin{align}
    I_{3}(k,j,l)&= \mathop{\EE}_{\alpha} g^\prime(\lambda^k)\lambda^k (\lambda^j) \\
    I_{32}(k,j)&= \mathop{\EE}_{\alpha} g^\prime(\lambda^k)\lambda^j \\
    I_{31}(k)&= \mathop{\EE}_{\alpha} g^\prime(\lambda^k)  \\
    I_{22}(k,j)&= \mathop{\EE}_{\alpha} g^\prime(\lambda^k) g(\lambda^j).
\end{align}
\end{subequations}

The update of $q$ can similarly be decomposed as a sum over the different Gaussian clusters:
\begin{align}
    \frac{\partial q^{kl}(\rho,t)}{\partial t} =\sum_{\alpha\in\mathcal{S}\left(+\right)} \mathcal{P}_\alpha \frac{\partial q_{\alpha^+}^{kl}(\rho,t)}{\partial t} +\sum_{\alpha\in\mathcal{S}\left(-\right)} \mathcal{P}_\alpha\frac{\partial q_{\alpha^-}^{kl}(\rho,t)}{\partial t}
\end{align}

The linear contribution to this update is directly computed by using Eq.~\eqref{eq:delta_wtau_alpha} and is similar to the one for $ m^{\beta k}(\rho,t)$. The quadratic contribution is obtained by using the fact that the projected inputs $\tilde x_\tau$ have a correlation of order $\nicefrac{1}{\sqD}$ with the local fields. Therefore, to leading order, this contribution is given by terms of the form:
\begin{align}
    \frac{\eta^2}{D^2}\sum_\tau\rho_\tau\mathop{\EE}_{\alpha}\left[g^\prime(\lambda^k)g^\prime(\lambda^l)g(\lambda^i)g(\lambda^j)\tilde x^2 _\tau\right] = \frac{\eta^2}{D} \mathop{\EE}_{\alpha}\left[g^\prime(\lambda^k)g^\prime(\lambda^l)g(\lambda^i)g(\lambda^j)\right] \left(\sum_\tau\rho_\tau \mathop{\EE}_{\alpha}\left[\tilde x^2 _\tau \right]\right)+O(D^{-3/2})
\end{align}
Let us define the constant of motion $\chi^\alpha=\frac 1 D \sum_\tau \rho_\tau^2 $, then the quadratic term in the update for $q_{\alpha^{\pm}}$ is given by:
\begin{align}
   \eta^2\chi^\alpha v^k v^l \left( I_{42}(k,l)\mp 2\sum_j v^j I_{43}(k,l,j) +\sum_{j a}v^j v^a I_{4}(k,l,j,a) \right).
\end{align}
The multidimensional integrals $I$ are given by:
\begin{equation*}
     I_{4}(k,l,j,a)=\mathop{\EE}_{\alpha} g'(\lambda^k)g'(\lambda^l)g(\lambda^j)g(\lambda^a) \quad I_{43}(k,l,j)=\mathop{\EE}_{\alpha} g'(\lambda^k)g'(\lambda^l)g(\lambda^j) \quad I_{42}(k,l)=\mathop{\EE}_{\alpha} g'(\lambda^k)g'(\lambda^l). 
\end{equation*}
Finally, the full equation of motion of $q$ is written:
\begin{align}
    \label{eq:eom-q}
    \begin{split}
   \frac{\partial q^{k l}(\rho,t)_{\alpha^\pm}}{\partial t}=&\pm \eta v^k I_{31}(k)m^{\alpha k}(\rho,t)\pm\frac{ \eta \rho  v^k}{Q^{kk}}\left( I_{32}(k,k)- I_{31}(k) M^{\alpha k} \right)q^{k l}(\rho, t) \\
   &+ \sum_{j\neq k}\left[ -\eta v^k v^j
   I_{22}(k,j)m^{\alpha k}(\rho, t) \right.\\
    &\qquad +\left.\frac{\eta \rho v^k v^j}{Q^{kk}Q^{jj}-Q^{kj2}}\left\{\left(-I_{3}(k,k,j)+ I_{32}(k,j) M^{\alpha k} \right)\left(Q^{jj}q^{k l}(\rho,t)-Q^{kj}q^{jl}(\rho,t)\right)\right.\right. \\
    &\hspace{11.5em}+\!\!\left.\left.\left(- I_3(k,j,j)+ I_{32}(k,j) M^{j\alpha} \right)\left(Q^{kk}q^{jl}(\rho,t)-Q^{kj}q^{kl}(\rho,t)\right)\right\}\right] \\
    & - \eta v^k v^k I_{22}(k,k)m^{\alpha k }(\rho,t)-\frac{\eta \rho v^k v^k}{Q^{kk}}\left( I_3(k,k,k)- I_{22}(k,k) M^{\alpha k} \right)q^{k l}(\rho,t) \\
    &+\left(k\leftrightarrow l\right) \\
    &+\eta^2\chi^\alpha v^k v^l \left( I_{42}(k,l)\mp 2\sum_j v^j I_{43}(k,l,j) +\sum_{j a}v^j v^a I_{4}(k,l,j,a) \right)-2\kappa\eta q^{k l}(\rho,t).
    \end{split}
\end{align}

\paragraph{Update for the second layer weights}

The update of the second layer weights is also decomposed into the contribution of the different Gaussian clusters and follows from taking the expectation of Eq.~\eqref{eq:sgdv} on the GM distribution~\eqref{eq:data}:
\begin{align}
\label{eq:eom-v}
\begin{split}
     \EE \frac{\dd v^k(t)}{\dd t} =&\sum_{\alpha\in\mathcal{S}\left(+\right)} \mathcal{P}_\alpha \frac{\dd v^k_{\alpha^+}(t)}{\dd t} +\sum_{\alpha\in\mathcal{S}\left(-\right)} \mathcal{P}_\alpha\frac{\dd v^k_{\alpha^-}(t)}{\dd t}, \\
    \frac{\dd v^k_{\alpha^\pm}(t)}{\dd t}=&\pm \eta \mathop{\EE}_{\alpha} g(\lambda^k) - \eta \sum_j v^j \mathop{\EE}_{\alpha} g(\lambda^k)g(\lambda^j)- \eta\kappa v^k
\end{split}
\end{align}

Equations \eqref{eq:eom-m},\eqref{eq:eom-q} and \eqref{eq:eom-v} suffice to fully characterise the training dynamics, in the limit of high dimensions and online-learning, of a 2LNN trained on an arbitrary Gaussian mixture with $O(1)$ clusters each having mean $\mu^\alpha$  and same covariance matrix $\Omega$. 

\paragraph{Agreement with Numerical Simulations}

\begin{figure*}[t!]
  \centering
  \includegraphics[width=\linewidth]{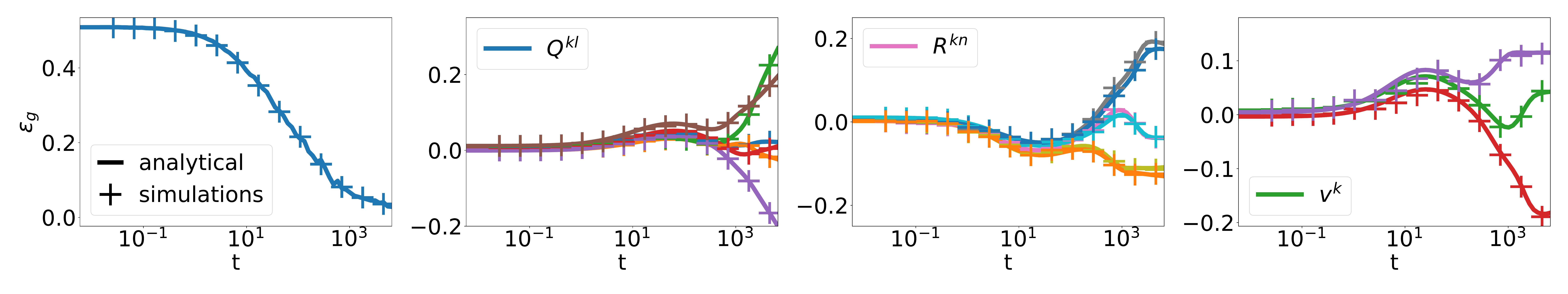}
  % \vspace{-0.5cm}
  \caption{\label{fig:ode-vs-sim} Agreement between simulations and ODEs when
    training a $K=3$ 2LNN on a Gaussian mixture in $D=800$ dimensions, with four
    Gaussian clusters with random covariance matrix $\Omega$ and random means
    (i.e. $\mu_r\!\sim\!\mathcal{N}(0,1)$) for sigmoidal activation function. We
    verify that even at finite ($D=800$) input dimension, the analytical
    prediction agree well with simulations. $\eta=0.1$,
    $g(x)=\erf\left(\nicefrac{x}{\sqrt{2}}\right)$, weights initialised with
    s.t.d. $\sigma_0=1$. Monte-Carlo integration performed with $10^{-4}$
    samples.}
\end{figure*}

Here, we verify the agreement of the ODEs derived above with simulation of 2LNN trained via online SGD. 

To start, Fig.~\ref{fig:ode-vs-sim} displays the dynamics of a $K=3$ network trained on a Gaussian mixture with 4 Gaussian
clusters having covariance matrix $\Omega= \nicefrac{F^T F}{\sqrt D}$ and means $\mu^\alpha$, where the elements of both the matrix $F\in\mathbb{R}^{D\times D}$ and the means are sampled i.i.d.~from a standard Gaussian distribution. The agreement between analytical prediction, given by integration of the ODEs, and simulations is very good both in the dynamics of the test error, of the order parameters and of the second layer weights. 

Note that the equations of motion describe the evolution of the densities $m$ and $q$ averaged over the input distribution. The agreement between this evolution and simulations justifies, at posteriori, the implicit assumption that the stochastic part of the SGD increment~\eqref{eq:sgd} can be neglected in the $D\to\infty$ limit. We can thus conjecture that in the $D\to\infty$ limit, the stochastic process defined by the SGD updates converges to a deterministic process parametrised by the continuous time variables $t\equiv\nicefrac{N}{D}$. We further add that the proof of this conjecture is not a straight-forward extension of the one of~\citet{goldt2019dynamics} for i.i.d.~inputs since here, one must take into account the density of the covariance matrix.

\begin{figure*}[t!]
  \centering
  \includegraphics[width=\linewidth]{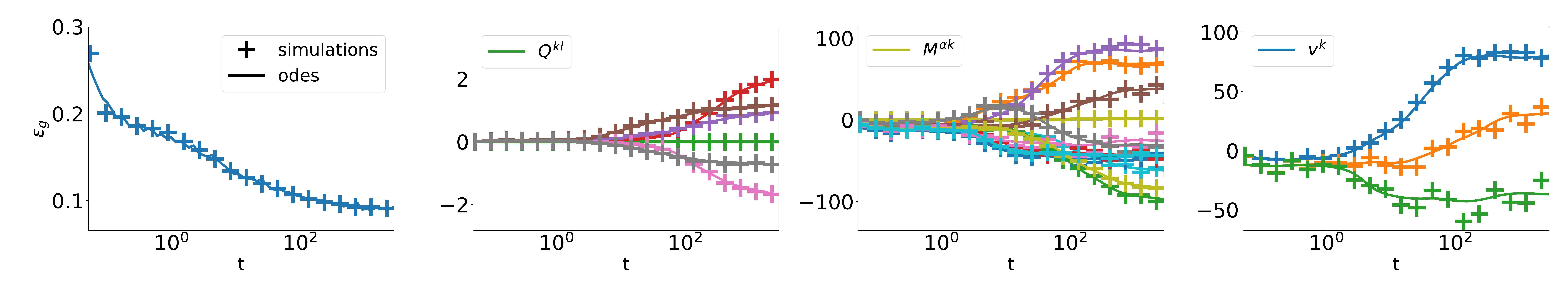}
  % \vspace{-0.5cm}
  \caption{\label{fig:ode-vs-sim_FMNIST} Agreement between simulations and ODEs when
    training a $K=3$ 2LNN with sigmoidal activation function on a Gaussian mixture obtained from the FashionMNIST dataset.
    The analytical dynamics obtained by integrating the ODEs agrees well with the given by simulations.
    Thus, the ODEs provide a tool to study the importance of datastructure in training 2LNN. We leave this study for future work. $D=784$, $\eta=0.1$,
    $g(x)=\erf\left(\nicefrac{x}{\sqrt{2}}\right)$, weights initialised with
    s.t.d. $\sigma_0=0.1$. Monte-Carlo integration performed with $10^4$
    samples.}
\end{figure*}

The ODEs are valid for generic covariance matrix and means. Thus, they can be used to analyse the role of data structure in training 2LNNs. Although we leave a detailed analysis for future work, Fig.~\ref{fig:ode-vs-sim_FMNIST} gives an example of how this could be done in the case where a $K=3$ 2LNN is trained on a GM obtained from the FashionMnist dataset.
The GM is obtained by computing the means $\bar{x}^\alpha$ and covariance matrix $\cov_{\alpha}(x,x)$ of each class in the dataset and assigning a label $+1$ or $-1$ to the different classes, as is commonly done in binary classification tasks. One could, for example, assign label $y=+1$ to the sneakers, boots, sandals, trousers and shorts categories and $y=-1$ to all others. Extending our analysis to $C$-class classification is straight forward and follows the analysis of~\citet{yoshida2019statistical}. The inputs are then sampled from a GM where the cluster's mean are given by $\bar{x}^\alpha$ and the covariance matrix $\Omega$ is the mean covariance of all classes: $\Omega=\nicefrac{1}{\text{n}_{\text{classes}}}\sum\limits_{\alpha} \cov_{\alpha}(x,x)$. Note the similarity between this procedure and \emph{linear discriminant analysis } commonly used in statistics. The agreement between simulations and analytical predictions is again very good, both at the level of the test error and of the order parameters.

\subsection{Simplified ansatz to solve the ODEs for the XOR-like mixture}
\label{app:ansatz-for-XOR}

Here, we detail the procedure, introduced in Sec.~\ref{sec:state-evolution}, used to find the long time $t\to\infty$ performance of 2LNN by making an \emph{ansatz} on the form of the order parameters that solve the fix point equations. The motivation for doing so, as argued in  of the main text, is that integrating the ODEs is numerically expensive as it requires evaluating various multidimensional integrals and the number of equations to integrate scales as $K^2$. 
In order to extract information about the asymptotic performances of the network, one can look for a fix point of the ODEs. However, the number of coupled equations to be solved, also scales quadraticaly with $K$ and is already $26$ for a $K=4$ student. The trick is to make an \emph{ansatz}, with fewer degrees of freedom, on the order parameters that solve the equations.
Used in this way, the ODEs have generated a wealth of analytical insights into
the dynamics and the performance of 2LNN in the classical teacher-student
setup~\cite{Biehl1995, Saad1995b, Saad1995b, Biehl1996, saad2009line,
  yoshida2019datadependence, yoshida2019statistical, goldt2020modelling}. In all
these works though, an important simplification occured because the means of the
local fields were all zero by construction. This simplification allowed the
fixed points to be found analytically in some cases. Here, the means of the
local fields evaluated over individual Gaussians in the mixture are \emph{not}
zero, so we have to resort to numerical means to find the fixed points of the
ODEs. 

Consider, for example, a 2LNN trained on the XOR-like mixture of Fig.~\ref{fig:snr_error}.
The Gaussian clusters have covariance
$\Omega=\sigma^2 \mathbb{I}$ and means chosen as in the left-hand-side diagram of Fig.~\ref{fig:snr_error}, with the remaining $D-2$ components set to 0. This configuration leads to the constrain
$w^k\cdot \mu^{\pm 0 }=-w^k\cdot \mu^{\pm 1 }$ that, in terms of overlap matrices, forces $M^{\alpha k}=-M^{\alpha+1 k}$ thus halving the number of free
parameters in $M$. 
It is also clear that the only components of the weight vectors which contribute to the error are those in the plane spanned by the means of the mixture. The additional $D-2$ components can be taken to $0$: i.e.~$w^k_r = 0$ for $r=2,..,D$. This condition allows to decompose the weight vectors as:
\begin{align}
    w_s^k = \sum_r \frac{w^k_r\mu_r^{+0}}{\sqD}\frac{\mu_s^{+0}}{\sqD} +  \sum_r \frac{w^k_r\mu_r^{-0}}{\sqD}\frac{\mu_s^{-0}}{\sqD}
\end{align}
This decomposition, fully constrains the overlap matrix $Q^{kl}$ in terms of $M^{\alpha k}$:
\begin{align}
\begin{split}
   Q^{kl} &= \sigma^2 \sum_s \frac{w_s^k w_s^l}{D}\\
   &=\sigma^2  \left(\sum_r \frac{w^k \mu_r^{+0}}{D}\right) \left(\sum_a \frac{w^l_a\mu_a^{+0}}{D}\right)\frac{|\mu^{+0}|^2}{D} + \sigma^2  \left(\sum_r \frac{w^k_r\mu_r^{-0}}{D}\right) \left(\sum_a \frac{w^l_a\mu_a^{-0}}{D}\right)\frac{|\mu^{-0}|^2}{D}\\
 \Longrightarrow Q^{kl} & = \sigma^2 \left( M^{+0 k } M^{+0 l } + M^{-0 k } M^{-0 l }\right),
\end{split}
\end{align}
where we used that in the XOR-like mixture, $|\mu^\alpha|=\sqD$ and $\mu^{0+}\cdot \mu^{0-} = 0$.
From the symmetry between the positive and negative sign clusters of the mixture, in the fix point configuration, for every weight having norm $|w|$ and at an angle $\alpha$ with the mean of a positive cluster, there is a corresponding weight of the same norm, at an angle $\alpha$ with a negative mean. I.e.~the angles of the weight vectors $w^k$ to the means $\mu^\alpha$, as well as the norms of the weights, are $2\times 2$ equal (one for the positive sign cluster and the other for the negative sign one). This constrains further half  the number of free parameters in the overlap matrix $M$, which are down to $4 K /(2 \times 2) = K$.
The second layer weights $v$ are fully constrained by requiring the output of the student to be $\pm 1$ when evaluated on the means. Putting everything together, one is left with $K$ equations to solve for the $\nicefrac{K}{2}$ angles and the $\nicefrac{K}{2}$ norms, or equivalently, for the $K$ free parameters in the overlap matrix $M$.
The agreement between the solution found by solving this reduced set of equations and simulations is displayed both in Fig.~\ref{fig:SE_var_reg}, where we use it to predict the evolution of the test error with the $L2-$regularisation constant.
\section{Transforming a Gaussian mixture with random features}
\label{app:rf}
\subsection{The distribution of random features is still a mixture}
Given an input $x = (x_i) \in \mathbb R^D$ sampled from the
distribution~\eqref{eq:data}, we consider the \emph{feature} vector
$z=(z_i)\in \mathbb R^N$
\begin{align}
\label{eq:feature}
    z_i=\psi(u_i), \qquad u_i \equiv \sum_{r=1}^D\ussD F_{ir}x_r
\end{align}
where $F\in  \mathbb{R}^{D\cross P}$ is a random projection matrix and
$\psi: \mathbb R\to \mathbb R$ is an element-wise non linearity. The
distribution of $z$ can be computed as:
\begin{align} p_z(z)&=\int_{\mathbb R^D}d x p_x( x) \delta\left(
z-\psi\left(\frac{ x F}{\sqrt D}\right)\right) \notag\\ &=\int_{\mathbb R^D}d x
\sum_{y} q(y) q( x | y) \delta\left( z -\psi\left(\frac{ x F}{\sqrt
D}\right)\right) \notag\\ &\equiv \sum_{y} q(y)
\sum_{\alpha\in\mathcal{S}\left(y\right)} \mathcal{P}_\alpha p_z^\alpha( z),
\end{align} with
\begin{equation} p_z^\alpha( z)=\int_{\mathbb R^D}d x \delta\left(
z-\psi\left(\frac{ x F}{\sqrt D}\right)\right)
\normal_\alpha\left(\frac{\mu^\alpha}{\sqrt{D}}, \Omega^\alpha\right)
\end{equation}
Crucially, the distribution of the features $z$ is still a
mixture of distributions.
We can thus restrict to studying the transformation of a Gaussian random variable
\begin{equation}
  x_r = \frac{ \mu_r}{\sqrt{D}} + \sigma w_r
\end{equation}
where $w_r$ is a standard Gaussian. The scaling of $\mu_r$ and $\sigma$ is chosen according to which regime (low or high $\snr$) one
chooses to study. We aim at computing the distribution, in particular the two first moments, of the feature $z$
defined in Eq.~\eqref{eq:feature}.  By construction, the random variables $u_i$
are Gaussian with first two moments:
\begin{align}
 \label{eq:u}
  \EE u_i = \frac{1}{\sqD} \tilde \mu_i, \qquad \tilde \mu_i \equiv \sum_r \frac{F_{ir} \mu_r}{\sqD}
\\
  \EE u_i u_j = \frac{1}{D} \tilde{\mu}_i \tilde{\mu}_j + \sigma^2  \sum_r \frac{F_{ir} F_{jr}}{D}
\end{align}

\subsection{Low signal-to-noise ratio}
\label{app:rf_low-snr}

Here we compute the statistics of the features, for general activation function, in the low signal to noise regime, for which $\nicefrac{|\mu|}{\sqrt{D} }\sim O(1)$ and $\sigma\sim O(1)$ so that the Gaussian clusters are a distance of order 1 away from the origin.

The mean of $z_i$ can be written as:
\begin{equation}
    \EE z_i= \EE \psi(u_i) = \EE \psi\left( \frac{ \tilde \mu_i}{\sqD} + \sigma \zeta \right),
\end{equation}
where $ \zeta\in\mathbb{R}$ is a standard Gaussian variable. In the scaling we work in, where $P$ and $D$ are send to infinity with their ratio fixed, and $\nicefrac{|\mu|}{\sqD}\sim O(1)$, $\tilde \mu_i /\sqrt{D}$ is of order $O(\nicefrac{1}{\sqrt{D}})$. Thus, the activation function $\psi$ can be expanded around $\sigma \zeta$:
\begin{align}
\label{app:mean_z}
    \EE z_i= \EE \psi\left( \sigma \zeta \right) +  \frac{\tilde \mu_i }{\sigma \sqrt D} \EE \zeta\psi\left( \sigma \zeta \right)+O\left(\frac{1}{D}\right).
\end{align}
where we used integration by part to find $\sigma\EE \partial\psi\left( \sigma \zeta \right) = \EE \zeta\psi\left( \sigma \zeta \right)$. 
%We note how $\partial\psi\left( \sigma \zeta \right)$ is well defined even for ReLU activation function since $\zeta$ is a continuous random variable and has probability $0$ to be at $0$.

For the covariance matrix, we separate the computation of the diagonal from the off-diagonal. Starting with the diagonal elements:
\begin{align}
\label{app:cov_diag_z}
\begin{split}
    \EE[z_i^2] =&\EE\left[\left(\psi\left( \sigma \zeta \right)+ \frac{\tilde \mu_i }{\sigma \sqrt D} \partial \psi\left( \sigma \zeta \right) \right)^2\right]\\
=&\EE\left[\psi\left( \sigma \zeta \right)^2\right] +  \frac{\tilde \mu_i }{\sigma \sqrt D} \EE \zeta\psi^2\left( \sigma \zeta \right)+O\left(\frac{1}{D}\right)
\end{split}
\end{align}
where, once gain, integration by parts was used to obtain $2\sigma \EE\left[\psi\left( \sigma \zeta \right) \partial \psi\left( \sigma \zeta \right) \right] = \EE \zeta\psi^2\left( \sigma \zeta \right)$.

In order to compute the off-diagonal elements, we note that different components of $u$ are weakly correlated since $\operatorname{Cov}(u_i,u_j)=\sigma^2\sum_{r}\nicefrac{F_{ir}F_{jr}}{D}\!\sim\!O(1/\sqrt{D})$. We can therefore apply formula Eq.~\eqref{eq:2pt} for weakly correlated variables:
\begin{align}
    \mathop{\EE}_{(z_i,z_j)} z_i z_j&=\mathop{\EE}_{(u_i,u_j)} \psi(u_i)\psi(u_j) \notag\\
    %&=\mathop{\EE}_{u_i} \psi(u_i) \mathop{\EE}_{u_j} \psi(u_j) + \frac{1}{\sigma^2}\sum_{r}\frac{F_{ir}F_{jr}}{D}\left(\mathop{\EE}_{u_i} u_i\psi(u_i) -\frac{\tilde \mu_i}{\sqrt P}\mathop{\EE}_{u_i} \psi(u_i)\right)\left(\mathop{\EE}_{u_j} u_j\psi(u_j) -\frac{\tilde \mu_j}{\sqrt P}\mathop{\EE}_{u_j} \psi(u_j)\right)\notag\\
    &=\mathop{\EE}_{u_i} \psi(u_i) \mathop{\EE}_{u_j}\psi(u_j) + \frac{1}{\sigma^2}\sum_{r}\frac{F_{ir}F_{jr}}{D}\mathop{\EE}_{u_i} u_i\psi(u_i)\mathop{\EE}_{u_j} u_j\psi(u_j)+O(\frac{1}{D})
\end{align}
where the averages are now over the one dimensional distributions of $u_i\!\sim\!\mathcal{N }\left(\frac{\tilde \mu_i}{\sqrt D},\sigma^2\right)$.
We can now replace in the above $u_i= \frac{ \tilde \mu_i}{\sqD} + \sigma \zeta $ and keep only leading order terms: 
\begin{align}
\begin{split}
   \cov(z_i,z_j)&=\sum_{r}\frac{F_{ir}F_{jr}}{\sigma^2 D} \left(\psi(\sigma \zeta+ \frac{\mu_i}{\sqD})(\sigma \zeta+ \frac{\mu_i}{\sqD})\right) \left(\psi(\sigma \zeta+ \frac{\mu_j}{\sqD})(\sigma \zeta+ \frac{\mu_j}{\sqD})\right)\\
   &= \sum_{r}\frac{F_{ir}F_{jr}}{\sigma^2  D} \left(\psi(\sigma \zeta)\sigma \zeta+ O(\frac 1 \sqD)\right)\left(\psi(\sigma \zeta)\sigma \zeta+ O(\frac 1 \sqD)\right)\\
   &=\EE[\zeta\psi(\sigma \zeta)]^2 \sum_{r}\frac{F_{ir}F_{jr}}{D} + O\left(\frac 1 D\right)\quad \text{, for }i\neq j.
   \end{split}
\end{align}
Thus yielding the final result:
\begin{align}
\label{app:cov_offd_z}
   \cov(z_i,z_j)=\EE[\zeta\psi(\sigma \zeta)]^2 \sum_{r}\frac{F_{ir}F_{jr}}{D}\quad \text{, for }i\neq j.
\end{align}
We define the constants $a$, $b$ and $c$ as in Eq.~\eqref{eq:abc} and $d$ as:
\begin{equation}
    a = \EE \psi\left( \sigma \zeta \right), 
    \quad b= \EE \zeta\psi\left( \sigma \zeta \right),
    \quad c^2= \EE\left[\psi\left( \sigma \zeta \right)^2\right], 
    \quad d ^2 = \EE \zeta\psi^2\left( \sigma \zeta \right)
\end{equation}
These definitions together with Eq.~\eqref{app:mean_z}, Eq.~\eqref{app:cov_diag_z} and Eq.~\eqref{app:cov_offd_z} lead to the statistics of Eq.~\eqref{eq:rf_low-snr_mean} and Eq.~\eqref{eq:rf_low-snr_cov}:
\begin{align}
 \EE z_i &= a  +  \frac{\tilde \mu_i }{\sigma \sqrt D} b \notag\\
    \cov(z_i,z_j)&=
    \begin{cases}
    c^2 - a^2 + \underbrace{  \frac{\tilde \mu_i }{\sigma \sqrt D} \left(d^2-2ab\right)}_{O(\frac{1}{\sqD})\text{: subleading if $c^2-a^2>0$}}+O(\frac{1}{D})\quad \text{, if } i =j \\
    b^2 \sum_{r}\frac{F_{ir}F_{jr}}{D}+ O(\frac{1}{D}) \quad \text{, if }i\neq j 
    \end{cases}
\end{align}
The above shows that, the transformation of the means is only linear and in the low $\snr$ regime, the XOR-like mixture of Fig.~\ref{fig:snr_error}, is transformed into a XOR-like mixture in feature space which cannot be learned by linear regression.
Note, that the performance of linear regression on the features is equivalent to its performance on inputs $\tilde z\in\mathbb{R}^P$ sampled from a Gaussian equivalent model defined as:
\begin{equation}
    \tilde z_i = \EE z_i + \sum_j \Omega^{1/2}_{ij}\zeta_j,
\end{equation}
where $\cov(z_i,z_j)\equiv\sum_k \Omega^{1/2}_{ik}\Omega^{1/2}_{jk}$ and $\zeta\in\mathbb{R}^P$ is a random vector with components sampled i.i.d.~from a standard Gaussian distribution.

\subsection{ReLU features}
\label{app:relu_features}

In the case of Relu activation function i.e.~$\psi(x)=\max(0,x)$,
the mean and the covariance of the features can be evaluated analytically for
all $\snr$ regimes.
The distribution of the features within each
cluster is given by a modified Gaussian: the probability mass of the Gaussian on
the negative real axis is concentrated at the origin while the distribution on
the positive axis is unchanged. 

In particular, the integral to obtain the mean of $z$ can be computed analytically and is given by:
\begin{equation}
   \label{eq:app:relu_muZ}
   \EE z_i= \EE \operatorname{ReLU}(u_i)= \sigma \left[ \frac{\tilde  \rho_i}{2} \left(1+\operatorname{erf}(\frac{\tilde  \rho_i}{\sqrt{2}})\right)+\frac{1}{\sqrt{2\pi}} e^{-\tilde  \rho_i^2/2} \right], 
\end{equation}
where we defined:
$$\tilde \rho_i\equiv \frac{\tilde{\mu}_i}{\sqrt{D}\sigma}=\frac{\sum_r F_{ir}\mu_r}{D \sigma}.$$
The covariance is once again computed by separating the diagonal terms from the off-diagonal ones. The integral to obtain the diagonal terms has an analytical expression found to be:
\begin{align}
    \label{app:eq:cov_relu_diag}
    \cov(z_i,z_i)\! =\! \EE[\operatorname{ReLU}(u_i)^2]\!-\!\EE[\operatorname{ReLU}(u_i)]^2\! =\! \sigma^2\left[
    \frac{\tilde{\rho}_i}{\sqrt{2\pi}}e^{-\tilde{\rho}_i^2/2} + \frac{1}{2}(\tilde{\rho}_i^2+1)(1+\operatorname{erf}( \frac{\tilde{\rho}_i}{\sqrt{2}}) ) - (\EE z_i)^2\right]
\end{align}
For the off-diagonal components, we again use that the covariance of the
different components of the $u_i$ is of order $\nicefrac{1}{\sqrt{D}}$ as
$\cov(u_i,u_j)=\sigma^2 \sum_r \nicefrac{F_{ir} F_{jr}}{D}$. Therefore, to evaluate $\EE[\operatorname{ReLU}(u_i)\operatorname{ReLU}(u_j)]$, we can use the result Eq.~\eqref{eq:2pt} for weakly correlated variables with $\epsilon M_{12}=\sigma^2 \sum_r \nicefrac{F_{ir} F_{jr}}{D}$. Then to leading order in $\frac{1}{D}$, one finds:
\begin{align}
    \begin{split}
  \operatorname{Cov}(z_i,z_j) &= \EE[\operatorname{ReLU}(u_i)\operatorname{ReLU}(u_j)] -\EE[\operatorname{ReLU}(u_i)]\EE[\operatorname{ReLU}(u_j)] \\
  &=  \sum_r \frac{F_{ir} F_{jr}}{\sigma^2 D} \EE[u_i \operatorname{ReLU}(u_i)]\EE[u_j \operatorname{ReLU}(u_j)]+O(\frac{1}{D})
    \end{split}
\end{align}
where the expectations above are over one dimensional distributions $u_i\sim\mathcal{N}\left(\sum_r\frac{F_{ir\mu_r}}{D},\sigma^2\right)$. The integrals have an analytical closed form expression, which yields the final result for the covariance:
\begin{align}
    \label{eq:app:relu_covZ}
    \begin{split}
    \operatorname{Cov}(z_i,z_j)=\begin{cases}
    \sigma^2\left[
    \frac{\tilde{\rho}_i}{\sqrt{2\pi}}e^{-\tilde{\rho}_i^2/2} + \frac{1}{2}(\tilde{\rho}_i^2+1)(1+\operatorname{erf}( \frac{\tilde{\rho}_i}{\sqrt{2}}) ) - (\EE z_i)^2\right] \quad \text{if }i=j,\\
  \frac{\sigma^2}{4} \frac{\sum_r F_{ir}F_{jr}}{D}\left( 1+ \operatorname{erf}( \frac{\tilde{\rho}_i}{\sqrt{2}}) \right)\left( 1+ \operatorname{erf}( \frac{\tilde{\rho}_j}{\sqrt{2}}) \right)+O(\frac{1}{D})
    \qquad \text{if }i\neq j
    \end{cases}
        \end{split}
\end{align}

%by noting that the different components of $u$ are weakly correlated %and using Eq.~\eqref{eq:2pt} for variables with non zero mean. Up to %corrections of order $O(\nicefrac{1}{D})$, one finds:

%It is easy to check that in the low $\snr$ regime where %$\tilde{\rho}_i\!\sim\! O(1/\sqrt{D})$, these expression reduce to the %ones of Eq.~\eqref{eq:rf_low-snr_mean} and %Eq.~\eqref{eq:rf_low-snr_cov}:
%\begin{align}
%    &\EE z_i=\sigma\left(\frac{1}{\sqrt{2\pi}}+\frac{ %\tilde{\rho}_i}{2\sqrt{D}}\right)+O(\frac{1}{D})  \\
%    &\operatorname{cov}(z_i,z_j)\equiv\Omega_{ij}=\begin{cases}\sigma^%2
%   \frac{(\pi -1)}{2 \pi }+O(\frac{1}{\sqrt{D}})
%    \quad\text{if }i=j,\\
%    \frac{\sigma^2}{4} \frac{\sum_r F_{ir}F_{jr}}{D} +O(\frac{1}{D})  %\quad \text{if }i\neq j
%    \end{cases}
%\end{align}

\subsection{Relation with the kernel}
As discussed in Sec.~\ref{sec:rf_performances} of the main text , the performances of kernel methods can be studied by using the convergence of RF to a kernel in the $\gamma\to\infty$ limit taken after the $D,P\to\infty$ limit~\cite{rahimi2008random}:
\begin{equation*}
  K(x, y)\!=\!\frac{1}{P}\!\sum_{i=1}^P
  \mathop{\EE}_F\!\left[\psi\left(\sum_{r=1}^D\frac{x_r F_{ir}}{\sqrt
      D}\right)\psi\left(\sum_{s=1}^D\frac{y_s F_{is}}{\sqrt D}\right)\right],
\end{equation*}

In the low $\snr$ regime where $\nicefrac{|\mu|}{\sqD}\sim O(1)$, the action of the kernel is essentially linear. The three constants $a$, $b$ and $c$, defined in Eq.~\eqref{eq:abc}, can be expressed equivalently in terms of the activation function $\psi$ or of the kernel. 
Consider $\omega_1$, $\omega_2\in\mathbb{R}$, two i.i.d.~standard Gaussian random variables, and denote by angle brackets $\langle \cdot \rangle$ the expectation over $w_1, w_2$. Then, by the definition $K(x,y)$ one has:
\begin{equation}
    \langle K(\sigma \omega_1,\sigma \omega_1)\rangle = \frac{1}{P} \sum_{i=1}^P \langle \psi(u_i)^2 \rangle = c^2
\end{equation}
where $u\in\mathbb{R}^P$ is the random vector whose moments are defined in Eq.~\eqref{eq:u} and we used the element wise convergence of $\nicefrac{\psi(\sum_r F_{ir}x_r)\psi(\sum_r F_{is}y_s)}{P}$ to its expected value~\cite{rahimi2008random}. 
Similarly, one has:
\begin{equation}
    \langle K(\sigma \omega_1,\sigma \omega_2)\rangle = \frac{1}{P}\sum_{i=1}^P \langle \psi(u_i) \rangle^2 = a^2
\end{equation}
Finally, for $b$ one has to perform a linear expansion of the kernel around the noise variable $\sigma \omega_{\substack{1\\2}}$:
\begin{align}
  \langle K(\sigma \omega_1+ \frac{\mu}{\sqD},
  \sigma \omega_2 +\frac{\mu}{\sqD})\rangle =&\frac{1}{P}\sum_{i=1}^P
  \langle \mathop{\EE}_F\left[\psi(u_{i1}+\sum_r\frac{\mu_{r}F_{ir}}{D})\rangle\langle\psi(u_{i2}+\sum_r\frac{\mu_{r}F_{ir}}{D})\right]\rangle\notag \\
  =&\frac{1}{P}\sum_{i=1}^P
\langle\psi(u_i)\rangle^2
  + \sum_{rs}\frac{\mu_r\mu_s}{D^2}\sum_{i=1}^P\mathop{\EE}_F\frac{F_{ir}F_{is}}{P}\langle\psi^\prime(u_i)\rangle^2\notag\\
  =& a^2 + \frac{1}{D\sigma^2}b^2,\notag\\
  \Longrightarrow  b^2 =& D \sigma^2\left( -a^2 +\langle K(\sigma \omega_1+ \frac{\mu}{\sqD},
  \sigma \omega_2 +\frac{\mu}{\sqD})\rangle \right)
\end{align}
These expressions allow to express the statistical properties of the features $z$, and to asses the performance of RF and kernel methods, directly in terms of the kernel without requiring the explicit form of the activation function.

For completeness, we give the analytical expression of the kernel corresponding to ReLU random features, i.e.~$\psi(x)=\max(0,x)$:
\begin{equation}
\label{eq:kernel_relu}
    K_{\text{ReLU}}(x,y)=\frac{|x||y|}{8\pi D} \left\{2|\sin(\theta)|+\cos(\theta)\left(\pi  +2 \operatorname{Arctan}\left(\frac{\cos(\theta)}{|\sin(\theta)|}\right)\right) \right\},
\end{equation}
where we defined the angle $\theta$ between the two vectors $x,y\in \mathbb{R}^D$ such that $|x|,|y|>0$:
\begin{equation}
    \theta = \frac{x\cdot y }{|x||y|}
\end{equation}
From Eq.~\eqref{eq:kernel_relu}, one sees that in case of ReLU activation function, the kernel is an angular kernel, i.e.~it depends on the angle between $x$ and $y$.
\section{Final test error of random features}
\label{app:rf_error}

\begin{figure}[t!]
  \centering
  \includegraphics[width=0.5\linewidth]{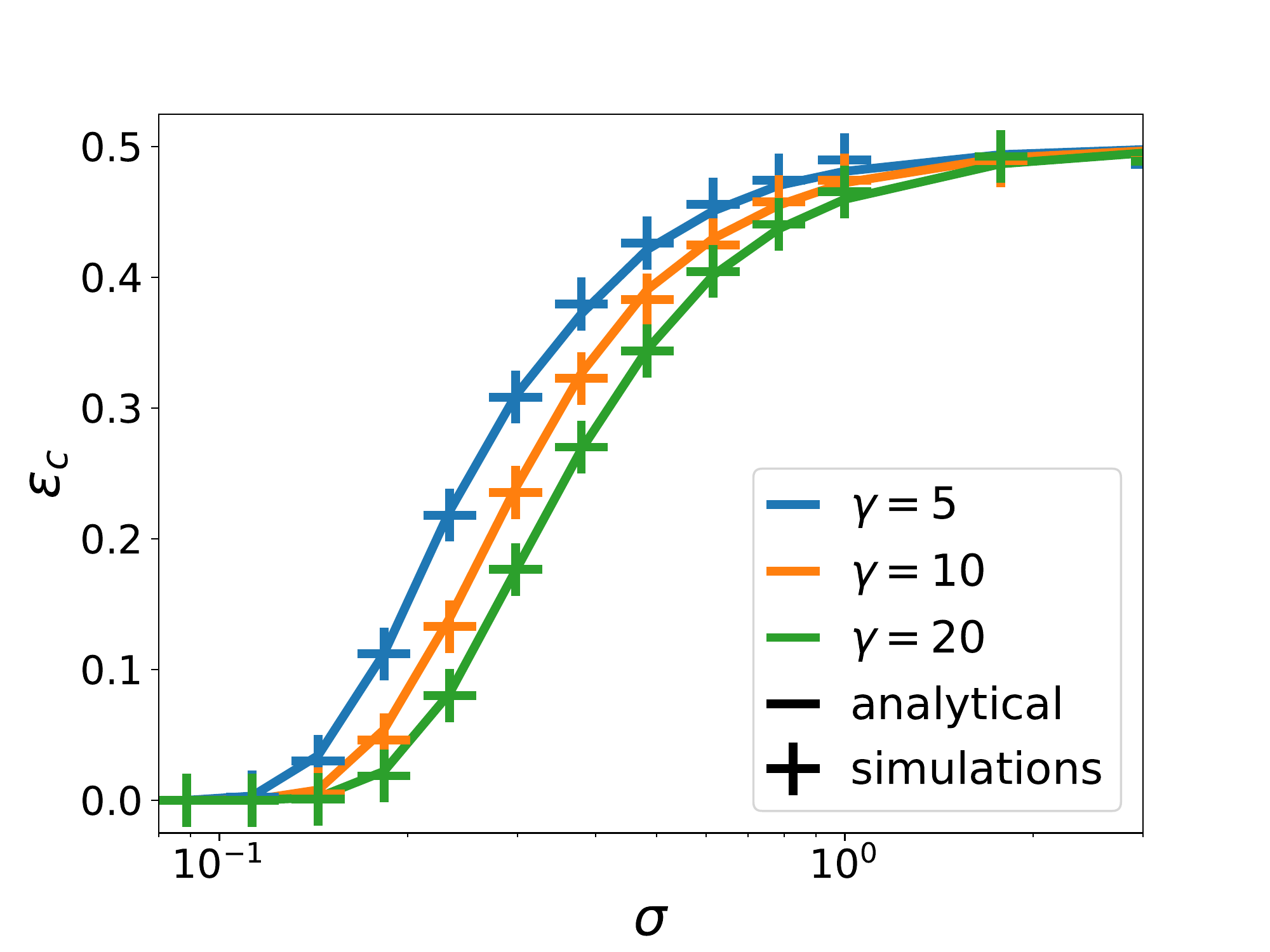}
  \caption{\label{fig:numerical_agreement_snr}
  Agreement between analytical predictions and simulations of the classification error of RF trained on the XOR like mixture of Fig.~\ref{fig:snr_error} for increasing $\sigma$ and various $\gamma$. The analytical predictions are obtained by computing the moments of the features $z$ using Eq.~\ref{eq:app:relu_muZ} and Eq.~\ref{eq:app:relu_covZ}. 
  These are then used to obtain, first the asymptotic solution of RF via Eq.~\ref{eq:asymptotic_rf_solution}, which is in turn plugged in Eq.~\ref{app:eq:MQz}. The final result is then given by Eq.~\ref{eq:classification_expectation_appendix}. Simulation results are obtained by training online an RF network with~$\eta=0.1$ until convergence. \emph{Parameters:}
  $D=800$, $\eta=0.1$, $P=\gamma D$, $\nicefrac{|\mu|}{\sqD}=1$.}
\end{figure}

This section details the computations leading to Eq.~\eqref{eq:rf_asymptotic_error} and Eq.~\eqref{eq:rf_asymptotic_classification_error} allowing to obtain the asymptotic performances of RF trained via online SGD on a mixture of Gaussian distribution.

Applying random features on $x\in\mathbb{R}^D$ sampled from the distribution~\eqref{eq:data} is equivalent to performing linear regression on the features
$z\in\mathbb{R}^P$ with covariance $\Omega^z$ and mean $\mu_z$. In the following, for clarity, we assume the features are centred, so that $\mu_z=0$, extending the computation to non centred features is straight-forward. The means of the individual clusters are however non zero. The output of
the network at a step $t$ in training is given by
$\phi(w)_t=\sum_{i=1}^P \nicefrac{w_i z_{ti}}{\sqP}$. We train the network in the
online limit by minimising the squared loss between the output of the network
and the label $y_t$. The $\pmse$ is given by:
\begin{align}
    \pmse(w)=\frac 1 2 \EE \left(y-\sum_{i=1}^P \frac{w_i z_i}{\sqP}\right)^2 = \frac 1 2 + \sum_{i,j=1}^P\frac{w_i w_j}{2 P}\Omega^z_{ij} - \sum_{i=1}^P\frac{w_i}{\sqP}\Phi_i,
\end{align}
where we introduced the \emph{input-label} covariance $\Phi_i\equiv\EE[z_i y]$.
  
The expectation of the SGD update over the distribution of $z$ is thus:
\begin{align}
  \label{eq:sgdRF}
  \EE \dd w_{i}&=-\frac{\eta}{\sqrt{P}} \EE z_i\left(\sum_{j=1}^P \frac{w_j z_j}{\sqP}-y\right)= \frac{\eta}{\sqrt{P}} ( \Phi_i - \sum_{j=1}^P\frac{w_j}{\sqP} \Omega^z_{ij})
\end{align}
Importantly, both the $\pmse$ and the average update only depend on the distribution of the features through the covariance matrix of the features $\Omega^z$ and the input label covariance $\Psi$.

To make progress, consider the eigen-decomposition of the covariance matrix $\Omega^z$:
\begin{equation}
  \Omega_{ij}^z = \frac{1}{P}\sum_\tau \rho_\tau \Gamma_{\tau i }\Gamma_{\tau j
  },
\end{equation}
where $\rho_\tau$ are the eigenvalues and $\Gamma_\tau$ their corresponding
eigenvectors.
%Note how, in the above, we write a sum over eigenvalues, since for numerical implementation, we always compute this decomposition at finite $D$.
%Extending the computation, to the $D\to\infty$ limit, where the sum is replaced by an integral over the bulk and a sum over the spikes of the spectrum of of $\Omega^z$ ..... DO WE NEED TO COMMENT ON THIS??

Define the rotation of $W$ and $\Phi$ into this
eigenbasis:
\begin{equation}
    \tilde w_\tau \equiv \frac{1}{\sqP}\sum_{i=1}^P \Gamma_{i\tau} w_i \quad
    \tilde \Phi_\tau \equiv \frac{1}{\sqP}\sum_{i=1}^P \Gamma_{i\tau} \Phi_i
\end{equation}
In this basis, the SGD update for the different components of $\tilde W$ decouple. One finds a recursive equation in which each mode evolves independently from the others:
\begin{align}
  \EE \dd \tilde w_{\tau}
  = \frac{\eta}{\sqrt{P}} ( \tilde \Phi_\tau - \frac{1}{\sqP}\sum_{\tau=1}^P\rho_\tau \tilde w_\tau)
\end{align}
Thus, the fix point $\tilde{\hat{W}} $ such that $\EE \dd \tilde{\hat{w}}_{\tau}=0$, can be found explicitly:  
$$ \rho_\tau \tilde{\hat{w}}_{\tau}=\sqP \tilde \Phi_\tau.$$ Rotating back in the original basis one finds the asymptotic solution for $W$ as:
\begin{equation}
\label{eq:asymptotic_rf_solution}
   \hat{w}_{i} = \sum_{\tau;\rho_\tau>0}  \frac{1}{\rho_\tau}\Gamma_{i\tau}\tilde \Phi_\tau. 
\end{equation}
The asymptotic test error is thus given by:

\begin{equation}
    \pmse_{t\to\infty} = \frac 1 2\left( 1 - \sum_\tau\frac{\tilde \Phi_\tau^2}{\rho_\tau} \right)
\end{equation}
 
\paragraph{Asymptotic classification error}
From the solution of Eq.~\eqref{eq:asymptotic_rf_solution} for the asymptotic solution found by linear regression, one can obtain the asymptotic classification error performed by random features as:
\begin{equation}
    \eclass{t\to\infty} = \EE \Theta(-y\lambda)= \sum_{\alpha}\mathcal{P}_\alpha \mathop{\EE}_\alpha \Theta(-y\lambda),
    \label{eq:classification_expectation_appendix}
\end{equation}
where $\Theta: \mathbb{R}\to\mathbb{R}$ is the Heaviside step function and we defined $\lambda\equiv\sum_{i=1}^P\nicefrac{\hat{w}_i  z_i}{\sqP}$.  
Introducing the local field $\lambda$ allows to transform the high dimensional integral over the features $z$ into a low-dimensional (in
this case one dimensional) expectation over the local field. The Gaussian
equivalency theorem of~\textcite{goldt2020gaussian} shows that even though the $z$
are not Gaussian, to leading order in $1/P$, the average Eq.~\eqref{eq:rf_asymptotic_classification_error},
only depends on the first two moments $\lambda$, defined as:
\begin{align}
\label{app:eq:MQz}
    M_\alpha = \mathop{\EE}_\alpha[\lambda] = \sum_{i=1}^P\frac{\hat{w}_i \mathop{\EE}_\alpha[z_i]}{\sqP}\quad 
    Q_\alpha = \mathop{\cov}_\alpha[\lambda,\lambda] = \sum_{i=1}^P\frac{\hat{w}_i\hat{w}_j}{P}\mathop{\cov}_\alpha(z,z)
\end{align}
These moments can be computed analytically from the statistics of the features computed in Sec.~\ref{app:rf} and from the optimal weights $W $ obtained in Eq.~\eqref{eq:asymptotic_rf_solution}. The classification error,
Eq.~\eqref{eq:classification_expectation_appendix}, can thus be evaluated by means of a one dimensional integral over the distribution of $\lambda$.

\begin{align}
\label{eq:rf_asymptotic_ec}
    \eclass{t\to\infty} &=  \sum_\alpha \mathcal{P}_\alpha\int \frac{d\lambda}{\sqrt{2 \pi Q_\alpha}}  \Theta(-y_\alpha\lambda) e^{-\frac{1}{2Q_\alpha} (\lambda-M_\alpha)^2} \\
    &=\frac{1}{2}\left(1-\sum_\alpha \mathcal{P}_\alpha y_\alpha  \erf\left(\frac{M_\alpha}{\sqrt{2 Q_\alpha}}\right) \right)
\end{align}

%The correctness of this result is demonstrated in Fig.~\ref{fig:snr_manygamma}, where we compare analytical predictions (lines) to simulations (crosses).
\section{The three-cluster model}
\label{app:three-clusters}

\begin{figure}[t!]
  \centering
  \includegraphics[width=\linewidth]{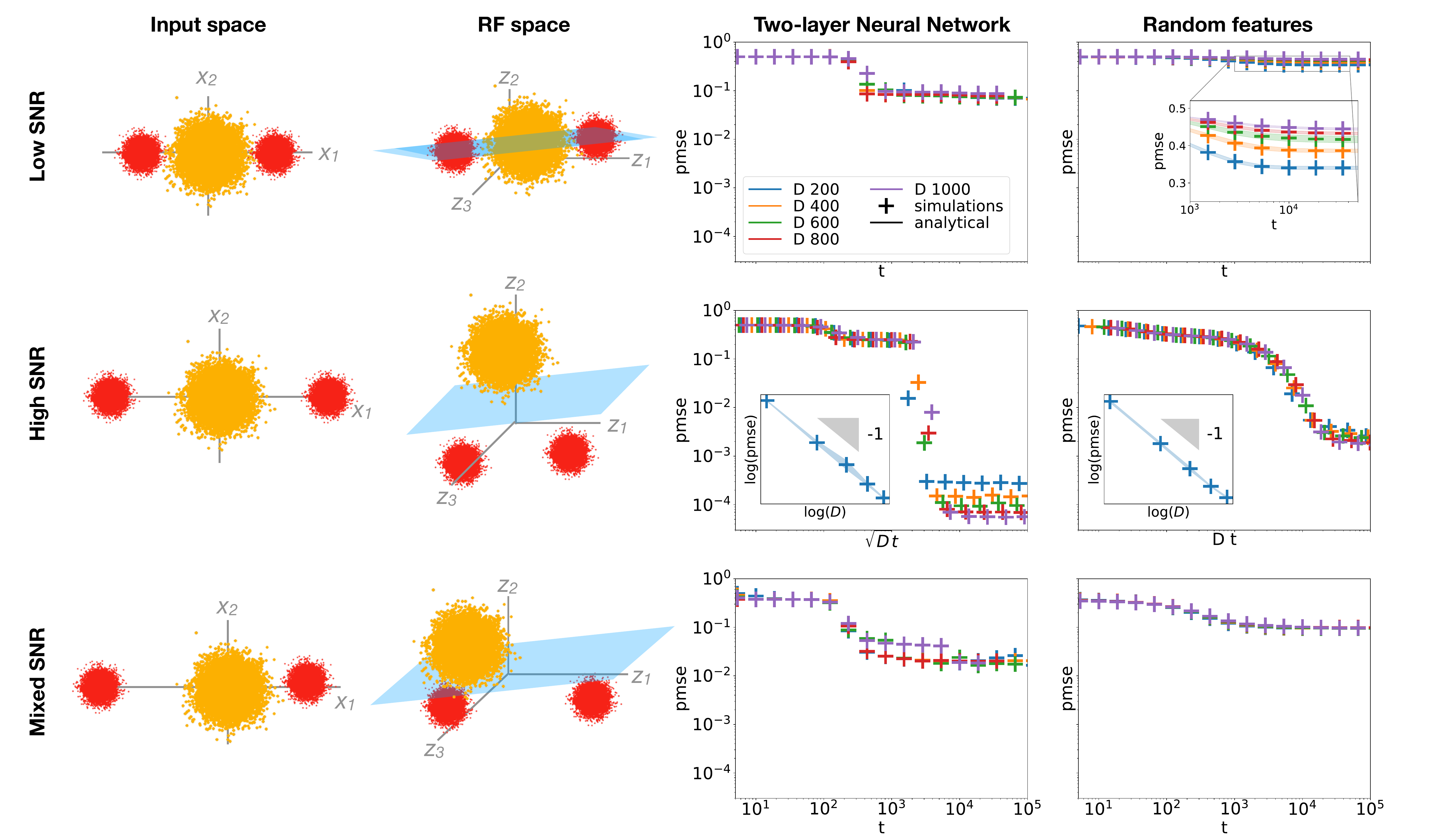}
  \caption{\label{fig:3cluster} We compare the performance of 2LNN with $K=10$ hidden units and ReLU activation function (\textbf{third} column) to the ones of random
      features (\textbf{fourth} column) on the three cluster problem with different signal-to-noise
      ratios. The \textbf{right} sketches the input space distribution and the \textbf{second} column the transformed distribution in feature space. 2LNN can considerably outperform RF in all three regimes. In the \emph{low} $\snr$ regime (\textbf{top}), the action of the random features is essentially linear inducing their performance to be as bad as random guessing in the $D\to\infty$ limit. In the \textit{high} $\snr$ regime (\textbf{middle}), instead, both networks manage to learn the task. In the \textit{mixed} $\snr$ regime (\textbf{bottom}), the distance between opposite sign clusters remains of orther one in feature space inducing the RF performances to be unchanged with $D$. We plot the test error as a function of time. \textit{Parameters:} $D=1000$, $\eta=0.1$, $\sigma^2=0.05$,
    $P=2D$ for random features, $K=10$ for 2LNN.}
\end{figure}

Similar to the analysis of the XOR-like mixture of Fig.~\ref{fig:xor}, we analyse a data model with three clusters that was the subject of several recent works~\cite{deng2019model, mai2019high, lelarge2019asymptotic, mignacco2020role, mignacco2020dynamical}. 
The Gaussian mixture in input space can be seen in the first column of Fig.~\ref{fig:3cluster}.
The means of both positive clusters are set to $0$ while the means of the negative sign clusters have first component $\pm \mu_0$ and all other $D-1$ components $0$. 
The mixture after random feature transformation is displayed in the second column and the third and fourth column show the performance of a 2LNN, respectively, a random feature network, trained via online SGD, on this problem. Here again, we build on the observation that overparametrisation does not impact performances and train a $K=10$ 2LNN in order to increase the number of runs that converged.
The three rows, are as before, three different $\snr$ regimes, they are in order the \textbf{low}, \textbf{high} and \textbf{mixed} $\snr$ regime.

The phenomenology observed in the XOR-like mixture carries through here. In the low $\snr$ regime, $\mu_0 = \sqD$ (\textbf{top row}), the 2LNN can learn the problem and its performance remains constant with increasing input dimension.
On the other hand, in this regime, the transformation performed by the random features is only linear in the large $D$ limit. Consequently, the RF performances degrade with increasing $D$ and are as bad as random chance in the limit of infinite input dimension.
In the \emph{high} $\snr$ regime instead (\textbf{second row}), where $\mu_0=D$, the mixtures becomes well separated in feature space allowing RF to perform well. Both the performance of 2LNN and RF improve as the clusters in the mixture become more separated. 
The \emph{mixed} $\snr$ regime (\textbf{bottom row}) is obtained by setting one of the negative sign clusters a distance $\sqD$ from the origin while maintaining the other one at a distance $1$. Here, the random feature perform a non trivial transformation of the far away cluster while its action on the nearby cluster is linear. Hence, in feature space, one of the negative clusters remains close to the positive clusters while the other is well separated. The RF thus achieve a test error which is better than random but still worse that that of the 2LNN. Its error is constant with increasing $D$ since it is dominated by the ``spill-over'' of the negative cluster into the positive cluster at the origin.

Lastly, let us comment that in all our work, we did not add a bias to the model. Adding a bias, does not change the conclusion that small 2LNN considerably outperform RF. In fact, the learning curves are only slightly modified. This is due to to our minimisation of the $\pmse$ when training the network, which, unlike classification loss that only cares about the \emph{sign} of the estimate, penalises large differences between label $y=\pm 1$ and the output. For simplicity, we thus chose to remove the bias in our analysis, although including it is a straight forward operation.

\end{document}